\documentclass[lettersize,journal]{IEEEtran}
\usepackage{amsmath,amsfonts}
\usepackage{array}
\usepackage{subcaption}
\usepackage{textcomp}
\usepackage{stfloats}
\usepackage{url}
\usepackage{verbatim}
\usepackage{multirow}
\usepackage{graphicx}
\usepackage{cite}
\usepackage{cite}
\usepackage{amsmath,amssymb,amsfonts}
\usepackage{amsthm}
\newtheorem{assumption}{Assumption}
\newtheorem{theorem}{Theorem}
\newtheorem{corollary}{Corollary}
\newtheorem{lemma}{Lemma}
\newtheorem{property}{Property}
\usepackage[linesnumbered,ruled,vlined]{algorithm2e}
\usepackage{mathtools}
\usepackage{graphicx}
\usepackage{textcomp}
\usepackage{xcolor}
\usepackage{bm}
\usepackage[
    colorlinks=true,    % enable colored links (no boxes)
    linkcolor=red,    % \ref, \eqref, \cref → black
    citecolor=blue,     % \cite → blue
    urlcolor=black,     % \url, \href → black
]{hyperref}

\DeclareMathOperator*{\argmax}{arg\,max}
\DeclareMathOperator*{\argmin}{arg\,min}

\usepackage{orcidlink}

\newcommand{\orcidSup}[1]{\textsuperscript{\orcidlink{#1}}}

\begin{document}

\title{Mobility-Assisted Decentralized Federated Learning: Convergence Analysis and A Data-Driven Approach}

\author{%
Reza Jahani\orcidSup{0009-0005-4289-2566}$^{1}$,
Md Farhamdur Reza\orcidSup{0000-0002-0377-8569}$^{1}$,
Richeng Jin\orcidSup{0000-0002-1480-585X}$^{2}$,
Huaiyu Dai\orcidSup{0000-0002-0078-4891}$^{1}$ \\
~\IEEEmembership{North Carolina State University$^{1}$, Zhejiang University$^{2}$}}

        % <-this % stops a space

% \markboth{}%


\maketitle

\begin{abstract}
Decentralized Federated Learning (DFL) has emerged as a privacy-preserving machine learning paradigm that enables collaborative training among users without relying on a central server. However, its performance often degrades significantly due to limited connectivity and data heterogeneity. As we move toward the next generation of wireless networks, mobility is increasingly embedded in many real-world applications. The user mobility, either natural or induced, enables clients to act as relays or bridges, thus enhancing information flow in sparse networks; however, its impact on DFL has been largely overlooked despite its potential. In this work, we systematically investigate the role of mobility in improving DFL performance. We first establish the convergence of DFL in sparse networks under user mobility and theoretically demonstrate that even random movement of a fraction of users can significantly boost performance. Building upon this insight, we propose a DFL framework that utilizes mobile users with induced mobility patterns, allowing them to exploit the knowledge of data distribution to determine their trajectories to enhance information propagation through the network. Through extensive experiments, we empirically confirm our theoretical findings, validate the superiority of our approach over baselines, and provide a comprehensive analysis of how various network parameters influence DFL performance in mobile networks.
\end{abstract}


\section{Introduction}
\noindent Federated learning (FL) has emerged as a privacy-preserving machine learning paradigm, enabling clients to collaboratively train a shared global model without sharing raw data~\cite{kairouz2021advances}. This algorithm has been widely used in many real-world applications, such as word-prediction~\cite{hard2018federated}, smart healthcare~\cite{kaissis2020secure}, etc. While conventional machine learning frameworks rely on a central server resulting in high network traffic and privacy concerns~\cite{chiang2016fog}, FL resolves these issues by introducing collaborative training. 
Despite the advantages that FL offers, it introduces several challenges, including security concerns \cite{zhang2022security} and communication bottlenecks \cite{wen2023survey}. Moreover, in real-world scenarios, data are often not independently and identically (non-IID) distributed, and FL suffers from significant performance degradation under heterogeneous data distribution across the clients \cite{zhao2018federated}. In these scenarios, conventional aggregation methods fail to achieve the desired performance~\cite{feng2021federated}.
% system

Numerous studies have advanced these directions by proposing new FL frameworks to address these challenges. Authors in~\cite{jin2024sign,jin2024ternaryvote} design FL frameworks to address security concerns, while other existing works focus on developing communication-efficient methods~\cite{elmahallawy2024communication,oh2024communication}. Many existing studies focus on addressing the challenges posed by data heterogeneity~\cite{wang2024aggregation, ye2023feddisco}, further improving the performance over conventional methods such as FedAvg and FedProx~\cite{mcmahan2017communication, li2020federated}.
Several other studies attempt to resolve the negative impact of data heterogeneity by adjusting the model at the local client side, with some focusing on incorporating regularization terms~\cite{acar2021federated} and some reducing the variance across optimized models~\cite{li2021model}, while other works proposing modifications at the server side~\cite{lin2020ensemble,zhang2022fine}.

% Numerous studies have advanced these directions by proposing new aggregation methods and addressing the challenges posed by data heterogeneity~\cite{wang2024aggregation, ye2023feddisco}, further improving the performance over conventional methods such as FedAvg and FedProx~\cite{mcmahan2017communication, li2020federated}. Authors in~\cite{jin2024sign,jin2024ternaryvote} design FL frameworks to address security concerns, while other existing works focus on developing communication-efficient methods~\cite{elmahallawy2024communication,oh2024communication}. 

In many settings, persistent connection to a central server for FL is unavailable or undesirable and reliance on a central server can introduce potential single point of failure~\cite{kairouz2021advances}.
Decentralized Federated Learning (DFL) removes the dependency on the central server, allowing the clients to exchange model updates through device-to-device (D2D) communication~\cite{beltran2023decentralized}.
This paradigm provides enhanced robustness and scalability while maintaining data privacy. However, it suffers from several fundamental challenges. Due to the absence of a global coordinator, the convergence of DFL algorithms heavily depend on the underlying network topology~\cite{zhu2022topology}, which determines how efficiently local updates are exchanged across the system~\cite{beltran2023decentralized}. Sparse and time-invariant topologies restrict information flow, leading to slower convergence or degraded accuracy, especially under non-IID data distributions across clients \cite{koloskova2020unified,bellet2022d,dandi2022data}.

Existing research has attempted to alleviate these issues by designing adaptive mixing matrices or heterogeneity-aware aggregation schemes~\cite{zhu2022topology,koloskova2020unified,dandi2022data,bellet2022d}. The authors in \cite{wu2024topology} also propose a topology learning algorithm for DFL under unreliable D2D networks. However, most of these approaches assume fixed connectivity graphs, overlooking the dynamic nature of wireless networks and embedded mobility.

With the proliferation of next generation of wireless networks, mobility is embedded in many applications. Several studies have considered mobility in server-based FL~\cite{feng2022mobility,peng2023tame} with some works unleashing the potential of FL with edge caching~\cite{yu2020mobility} and some works enabling wireless hierarchical FL~\cite{zhang2024ultra}. Additional research has proposed resource allocation and user scheduling strategies to address mobility in mobile networks~\cite{fan2025mobility}. However, the impact of client movement on fully decentralized systems remains largely unexplored.

A few recent studies have started to examine mobility in decentralized settings and address the challenges raised in DFL due to user mobility. In~\cite{chen2025mobility}, a leader selection strategy in vehicular networks is proposed to resolve resource constraints associated with vehicle mobility, further improving the DFL training efficiency. The authors in~\cite{wang2025decentralized} also address the challenges caused by sporadic connection of clients in a decentralized setting by proposing a cached-DFL framework. However, existing studies fail to account for network sparsity and limited connectivity among the clients under heterogeneous data distribution, leaving the impact of mobility on DFL systems in sparse networks not yet well understood. The author in~\cite{de2024exploring} suggests that mobility can dynamically alter the communication graph, further compensating for sparsity and accelerating convergence; however, it remains limited to random mobility and less practical settings.

Motivated by this research gap, we aim to systematically study the impact of mobility on DFL systems and design a robust DFL framework. We first derive a novel convergence bound for DFL under mobility, showing that the introduction of mobile clients enforces the \emph{B-strong connectivity} property in sparse networks. Building upon this theoretical insight, we propose \emph{Mobility-Assisted DFL} frameworks in which mobile clients follow distribution-aware trajectories, intelligently navigating the network to enhance information transfer among heterogeneous regions. Our mobility strategies, namely \emph{Distribution-Aware Movement (DAM)} and \emph{Distribution-Aware Cluster Center Movement (DCM)}, enable clients to exploit class-wise data distribution information and achieve efficient and scalable learning.

Our main contributions are summarized as follows:
\begin{itemize}
    \item We establish the convergence of DFL under user mobility, based on which two mobility strategies that leverage data distribution awareness are proposed to mitigate data heterogeneity in DFL.
    \item Through extensive experiments on MNIST and CIFAR-10, we confirm our theoretical insights and validate the superiority of our proposed framework over the baselines in various network configurations.
    \item We provide a comprehensive analysis of the impact of network parameters on DFL performance, further identifying trade-offs between deployment cost and system performance.
\end{itemize}

\section{System Architecture}\label{sec:sys}

\subsection{Network Setup}\label{sec:network_set}

\noindent We consider a decentralized network of $N$ clients denoted by the set $\mathcal{C} := \{1,..., N\}$. 
Among these clients, a subset $\mathcal{C}_m \subset \mathcal{C}$ comprises mobile clients that move according to a predefined mobility pattern, while the remaining clients, denoted as $\mathcal{C}_s = \mathcal{C} \setminus \mathcal{C}_m$, remain static throughout the process.
Each client $i \in \mathcal{C}$ holds a local dataset $\mathcal{D}_i= \{(\mathbf{s_n} , y_n) \mid  n=1,2,...,|\mathcal{D}_i|\}$ and updates its local model $\mathbf{x}_i$ based on its local dataset, where $\mathbf{s_n}$ and $y_n$ denote the $n$-th input and the corresponding true label. 
The DFL objective is to optimize the following function~\cite{koloskova2020unified}: \vspace{-2mm}
\begin{equation}
    f^* := \Big[\min_{\bm x \in \mathbb{R}^{dim}} f(\mathbf{x}) := \frac{1}{N} \sum_{i \in \mathcal{C}} f_i(\mathbf{x}) \Big],  \vspace{-2mm}
\end{equation}
where $f_i(\mathbf{x}) := \mathbb{E}_{\xi_i \sim \mathcal{D}_i} F_i(\mathbf{x}; \xi_i)$ represents the local expected loss over mini-batches $\xi_i$, with $F_i(.)$ denoting the client-specific loss function, and $d$ being the model dimensionality. 

We consider clients positioned on a $G \times G$ grid, where each location is represented as $ (p,q) \in \mathcal{G}$ with $\mathcal{G} = \{(p,q) \mid p,q \in \{1,2,...,G\}\}$. 
At the beginning of the training process, each client $i$ is randomly assigned an initial location $L_i^{(0)} \in \mathcal{G}$ within the grid. Each client $i$ has a limited communication range defined by a circular coverage area of radius $R_c$ and can exchange updates with clients located within this coverage. 
% A client $i$ can communicate and exchange updates with another client $j \in \mathcal{C}\setminus \{i\}$ if $j$ lies within this coverage.
The set of connected neighbors for client $i$ at global round $t$ is given by: $\mathcal{N}_i^{(t)} = \{j \mid \| L_i^{(t)} - L_j^{(t)}\|_2 \leq R_c, \forall j \in \mathcal{C}\setminus\{i\}\}$, where $L_i^{(t)}$ represents the location of client $i$ at the global round $t$. 
While fixed clients remain at the same locations, i.e., $L_i^{(t+1)} = L_i^{(t)}$, $\forall i \in \mathcal{C}_s$, mobile clients relocate to new locations within the grid according to a mobility pattern constrained by the maximum movable radius $R_m$, i.e. $\|L_{i_m}^{(t+1)}- L_{i_m}^{(t)}\| \leq R_m$ where $i_m \in \mathcal{C}_m$.
Consequently, the network topology varies over time due to clients mobility. 

\subsection{DFL System Model}\label{sec:dfl_sys}

\noindent In this decentralized framework, each client maintains local parameters $\mathbf{x}^{(t)}_i \in \mathbb{R}^{dim}$ and computes the local stochastic gradient $\mathbf{g}^{(t)}_i:= \nabla F_i(\mathbf{x}^{(t)}_i, \xi_i)$ based on samples from its dataset at round $t$. 
Following this, each client $i \in \mathcal{C}$ exchanges models with its neighbors $\mathcal{N}^{(t)}_i$ through D2D communication and aggregates the received models. The decentralized training process involves two phases per round:
\begin{enumerate}
    \item \textit{Local update}: $\mathbf{x}_i^{(t+\frac{1}{2})} = \mathbf{x}_i^{(t)} - \eta \mathbf{g}_i^{(t)}$, where $\eta$ is the learning rate.
    \item \textit{Consensus update}: $\mathbf{x}_i^{(t+1)} = \sum_{j=1}^N w^{(t)}_{j,i} \mathbf{x}_j^{(t+\frac{1}{2})}$, where $w^{(t)}_{j,i}$ is the $(j,i)$-th entry of the mixing matrix $\mathbf{W}^{(t)} \in \mathbb{R}^{N \times N}$, satisfying $w^{(t)}_{j,i}=0$ if $j \notin \mathcal{N}^{(t)}_i \cup \{i\}$.
\end{enumerate}
% For convergence, the mixing matrix $\bm{W}$ must be symmetric ($\bm W = \bm W^T$), doubly stochastic ($\bm W \bm 1 = \bm 1$ and $\bm 1^T \bm W = \bm 1^T$), where $\bm{1}$ is a vector of ones in $\mathbb{R}^C$. 
For convenience, we use the following matrix notation to stack all the models, 
$
    \mathbf{X}^{(t)} = \big[ \mathbf{x}_1^{(t)},..., \mathbf{x}_N^{(t)} \big] \in \mathbb{R}^{dim \times N}.
$
Likewise, we define $\mathbf{G}^{(t)} = [\mathbf{g}_1^{(t)},..., \mathbf{g}_N^{(t)}]$. Hence, the local and consensus updates for all the models can be expressed as: $\mathbf{X}^{(t+\frac{1}{2})} = \mathbf{X}^{(t)} - \eta \mathbf{G}^{(t)}$ and $\mathbf{X}^{(t+1)} = \mathbf{X}^{(t+\frac{1}{2})} \mathbf{W}^{(t)} $, respectively, where $\mathbf{W}^{(t)}$ is the doubly stochastic time-varying mixing matrix due to the mobile clients, i.e., $\mathbf{W}^{(t)}\bm1=\bm1$ and $\bm1^T \mathbf{W}^{(t)}=\bm 1^T$. 
Non-zero weights of the consensus matrix $\mathbf{W}^{(t)}$ are determined based on the degree of the nodes, based on the Metropolis-Hasting algorithm \cite{mynuddin2024decentralized}. The $(j,i)$-th entry of $\mathbf{W}^{(t)}$ is as follows: \vspace{-2mm}
\begin{equation}
w_{j,i}^{(t)} = 
    \begin{cases}
        \frac{1}{1+max\{deg_i^{(t)},deg_j^{(t)}\}}, & \text{if  }\; j \in \mathcal{N}_i^{(t)} ,\\
        1-\sum_{k \in \mathcal{N}_i^{(t)}}w_{k,i}^{(t)}, & \text{if }j=i,        \\
        0 , & \text{otherwise},
    \end{cases}
\end{equation}
where $deg_i^{(t)}$ and $deg_j^{(t)}$ correspond to the degree of node $i$ and $j$.
Algorithm~\ref{alg:dfl} illustrates how the DFL system works by local updates and aggregation when mobility is utilized in the network. This algorithm operates with the clustering of static clients based on their location ($\mathcal{L}$) and the mobility pattern under which the trajectory of mobile clients is controlled. We explicitly discuss the \textbf{ClientMobility} and \textbf{Clustering} algorithms in section~\ref{sec:framework} after we introduce baseline mobility models and our proposed mobility strategies.

\begin{algorithm}[h] \small
\caption{DFL Algorithm}
\label{alg:dfl}
\KwIn{Number of rounds $T$; Set of clients $\mathcal{C}$; Learning rate $\eta_t$; Clients location $\mathcal{L}$, Static clients location $\mathcal{L}_s$; Mobility pattern ($MV$), Grid locations $\mathcal{G}$, Communication radius $R_c$, Mobility Constraint $R_m$.}
\KwOut{Final trained local models $\{\mathbf{x}_k^{(T)}\}_{k\in\mathcal{C}}$}

\tcp{Time Complexity: $\mathcal{O}(T(N\cdot Cost_{ClientUpdate} + N\cdot dim + |\mathcal{C}_m|\cdot Cost_{ClientMobility}))$}

\If{$MV=$ DCM}{
    $\mathcal{L}_c =\text{ Clustering}(R_c,\mathcal{G}, \mathcal{C}_s, \mathcal{L}_s)$ \\
}
\ForPar{$k \in C$}{
        $\mathbf{x}_k^{(0)} \gets \mathbf{x}^{(0)}$\
    }
\For{$t = 0,1, 2, \dots, T-1$}{
    \ForPar{$k \in C$}{
        $\mathbf{x}_k^{(t+\tfrac{1}{2})} \gets \text{ClientUpdate}(k, t, \mathbf{x}_k^{(t)},\eta_t)$\
    }
    \ForPar{$k \in C$}{
        % $\omega_{ik}^{(t)} \gets [\bm W^{(t)}]_{ik}$\;
        $\mathbf{x}_k^{(t+1)} \gets \sum_{i \in \mathcal{N}_k^{(t)}} \omega_{ik}^{(t)} \mathbf{x}_i^{(t+\tfrac{1}{2})}$\
    }
    \ForPar{$i_m \in C_m$}{
        % $L_{i_m}^{(t+1)} \gets \text{ClientMobility}(i_m,R_m, L_{i_m}^{(t)},MV)$\;
        $\mathcal{L}[i_m] \gets \text{ClientMobility}(i_m,L_{i_m}^{(t)},R_m,\mathcal{C}_s,\mathcal{G},\mathcal{L}_c,MV)$ \\
        % $\mathcal{L}[i_m] \gets L_{i_m}^{(t+1)}$\
    }
    $\mathbf{W}^{(t+1)} \gets \mathcal{W}(\mathcal{L}, R_c)$\
}
\end{algorithm}

\begin{algorithm} \small
\caption{ClientUpdate$(k, t, \mathbf{x}_k^{(t)},\eta_t)$}
\label{alg:clientupdate}
\KwIn{Client $k$; round $t$; local model $\mathbf{x}_k^{(t)}$; learning rate $\eta_t$}
\KwOut{Updated Model $\mathbf{x}_k^{(t+\tfrac{1}{2})}$}
\tcp{Time Complexity: $\mathcal{O}(|\xi_k^{(t)}| \cdot dim)$}
Sample mini-batch $\xi_k^{(t)} \sim \mathcal{D}_k$\ \\
Compute gradient $\mathbf{g}_k^{(t)} \gets \nabla F_k(\mathbf{x}_k^{(t)}, \xi_k^{(t)})$\ \\
Update local model: $\mathbf{x}_k^{(t+\tfrac{1}{2})} \gets \mathbf{x}_k^{(t)} - \eta_t \mathbf{g}_k^{(t)}$\
\end{algorithm}

\subsection{Baseline Mobility Models}\label{sec:baseline}

\noindent In this section, we elaborate on the baseline mobility patterns in DFL systems. 
\subsubsection{Static}

For this scenario, all clients remain in their initial locations throughout the training procedure. Formally, $L_i^{(t+1)} = L_i^{(t)}; \forall i \in \mathcal{C}, \forall t \in [T]$, where $T$ is the total number of global training rounds. In other words, all clients are static and there is no mobile client, i.e. $|\mathcal{C}_m|=0$.

\subsubsection{Random Movement}
For this mobility pattern, mobile clients $i_m \in \mathcal{C}_m$ are allowed to change their location throughout the training procedure. These clients randomly select a new location $L_{i_m}^{(t+1)}$ within the grid based on a uniform distribution among the grid points in the network. This movement is subject to the mobility constraint $R_m$, i.e. maximum allowable displacement per round. This constraint ensures a feasible and practical range of movement for mobile clients. Formally,

\vspace{-5mm}

\begin{equation}
    L_{i_m}^{(t+1)} \sim \text{Uniform}\Big(\{L \mid \|L-L_{i_m}^{(t)}\|_2 \leq R_m, L \in \mathcal{G} \}\Big).
\end{equation}

\subsubsection{Connectivity-Oriented Movement}
To adopt a stronger baseline, we introduce Connectivity-Oriented Movement (COM), which enhances network connectivity through mobility. In this strategy, each mobile client evaluates all locations in $\mathcal{G}$ and assigns a probability to each grid point proportional to its connectivity (measured by the corresponding node degrees), subsequently selecting the next destination accordingly. 

To this end, the mobile client $i_m$ constructs the vector $\Delta\mathbf{d}_{i_m}^{(t)} = [\Delta deg_{i_m}^{(L,t)}]_{L \in \mathcal{G}}$, where $\Delta deg_{i_m}^{(L,t)} = deg_L - deg_{i_m}^{(t)}$, with 
$deg_L = |\{ j \mid \|L_j^{(0)} - L \|_2 \leq R_c, \, j \in \mathcal{C}_s \}|$ 
denoting the number of static neighbors after moving to the location $L$, and 
$deg_{i_m}^{(t)} = |\mathcal{N}_{i_m}^{(t)} \setminus \mathcal{C}_m|$ 
representing the current number of static neighbors. Each location in the network is assigned a probability by applying softmax\footnote{The softmax function can generate these probabilities, favoring locations with more neighbors compared to the current location.} to $\Delta\mathbf{d}_{i_m}^{(t)}$, according to which the mobile client selects a destination and moves toward that point subject to the mobility constraint $R_m$.
The workflow of this baseline mobility pattern can also be found in Algorithm \ref{alg:mobility} that we discuss in Section \ref{sec:framework}.

\begin{table}[!t]
\caption{Notations Definitions\label{tab:deftab}}
\centering
\begin{tabular}{|c||c|}
\hline
$\mathcal{C}$ & Set of clients\\
\hline
$\mathcal{C}_m$ & Set of mobile clients\\
\hline
$\mathcal{C}_s$ & Set of static clients\\
\hline
$\mathbf{x}_i$ & Client $i$ model \\
\hline
$\mathbf{X}$ & Concatenated clients models \\
\hline
$dim$ & Client parameter dimension \\
\hline
$\mathbf{G}$ & Concatenated clients gradients \\
\hline
$\mathcal{D}_i$ & Dataset of client $i$\\
\hline
$\mathbf{s}_n, y_n$ & n-th input and ground truth \\
\hline
$f_i(.)$ & Local objective function of client $i$ \\
\hline
$\xi_i^{(t)}$ & Client $i$ mini-batch at time $t$ \\
\hline
$F_i(\mathbf{x}_i^{(t)};\xi_i^{(t)})$ & Loss function over mini-batch for client $i$ \\
\hline
$R_c$ & Communication Radius \\
\hline
$\mathcal{L}$ & Set of clients locations \\
\hline
$\mathcal{L}_s$ & Set of static clients locations \\
\hline
$\mathcal{L}_c$ & Set of cluster center locations \\
\hline
$L_i^{(t)}$ & Client $i$ location at round $t$ \\
\hline
$\mathcal{N}_i^{(t)}$ & Client $i$ neighborhood at round $t$ \\
\hline
$R_m$ & Mobility Constraint \\
\hline
$\mathbf{g}_i^{(t)}$ & Gradient of mini-batch loss function \\
\hline
$w_{j,i}$ & Communication weight of client $j\rightarrow i$ \\
\hline
$\textbf{W}^{(t)}$ & Mixing Matrix including all $\omega_{ij}$ \\
\hline
$\eta$ & Learning Rate \\
\hline
$\mathcal{G}$ & Set of all grid locations in network \\
\hline
$[Y]$ & Set of all classes, $\{0, 1, 2, ..., Y-1\}$ \\
\hline
$G$ & Size of the square shaped network \\
% \hline
% $|\mathcal{C}_m|$ & Number of mobile nodes \\
\hline
$\alpha$ & Data heterogeneity level in experiments \\
\hline
$[T]$ & Set of global rounds, $[T]=\{0,1,2,...,T-1\}$ \\
\hline
$\mathcal{G}_g$ & Graph representing the connection among clients \\
\hline
$\mathcal{V}$ & Set of nodes in graph $\mathcal{G}_g$ \\
\hline
$E$ & Set of edges of graph $\mathcal{G}_g$ \\
\hline
$deg_L$ & Degree of a mobile client client at location $L$\\
\hline
$deg_{i_m}^{(t)}$ & Degree of mobile client at round $t$\\
\hline

\end{tabular}
\end{table}

\vspace{-3mm}

\section{Theoretical Analysis}\label{sec:theory}

\noindent In this section, we investigate the convergence of DFL under sparse network settings where clients are only allowed to communicate with neighboring clients restricted to communication radius $R_c$. 
% This setting differs from most of the existing studies in the literature \cite{wu2024topology, chen2025mobility, wang2025decentralized, de2024exploring}, 
In this setting, DFL convergence is not theoretically guaranteed under the static scenario due to the sparsity level of the mixing matrix. 

We first provide a high-level illustration of this observation, showing the persistence of sparsity in the network when there is no mobile client. We then extend our illustration to mobile networks and show that in the presence of mobile clients, the communication graph would satisfy the \textbf{B-Strongly Connected} property, which we will discuss further. Based on this property, we provide a representation for the models based on the models and gradients in the previous $B$ time steps. 
Explicitly, we derive the relationship between the models in each time step with the models and gradients in the previous $B$ rounds. We will further use that representation to establish the final convergence bound under mobility. 
% Motivated by the insight from the representation, we design our systematic DFL framework empowered by mobility strategies.

\noindent\textbf{Assumptions}. Our analysis is based on the following assumptions in DFL\cite{le2023refined}. 

\begin{assumption}[Smoothness]\label{assump:smooth} 
There exists a constant $L>0$ such that for any $\xi\in D_i$, $\mathbf{x},\mathbf{y}\in \mathbb{R}^{dim}$ we have. $\forall \ i\in\mathcal{C}; \|\nabla F_i(\mathbf{x},\xi )-\nabla F_i(\mathbf{y},\xi)\| \leq L \|\mathbf{x}- \mathbf{y}\|$.    
\end{assumption}

\begin{assumption}
[Bounded Variance]\label{assump:var} 
For any node $i \in \mathcal{C}$, there exists a constant $\sigma_i^2$ such that for any $\mathbf{x}\in \mathbb{R}^{dim}$, we have 
$\mathbb{E}_{\xi\sim D_i} \|\nabla F_i(\mathbf{x},\xi) - \nabla f_i(\mathbf{x}) \| \leq \sigma_i^2$.    
\end{assumption}

\begin{assumption}
[Mixing Parameter]\label{assump:mixing} There exists a mixing parameter $p \in [0,1]$ such that for any matrix $\mathbf{M}\in \mathbb{R}^{dim\times N}$, for any $t\in[T]$, we have
$\|\mathbf{M} \mathbf{W}^{(t)}-\bar{\mathbf{M}} \|_F^2 \leq (1-p)\|\mathbf{M}-\bar{\mathbf{M}} \|_F^2$, where $\|.\|_F$ denotes the Frobenius norm and $\bar{\mathbf{M}}=\mathbf{M}(\frac{\bm 1\bm1^T}{N})$.
\end{assumption}

Assumption \ref{assump:mixing} indicates how close an aggregation step with matrix $\mathbf{W}^{(t)}$ in round $t$ can bring an arbitrary matrix $\mathbf{M}$ closer to $\bar{\mathbf{M}}$. It is always validated for $p=1-\lambda_2(\mathbf{W}^{(t)^{T}}\mathbf{W}^{(t)})$ given that $\lambda_2(\mathbf{W}^{(t)^{T}}\mathbf{W}^{(t)})$ denotes the second largest eigenvalue of $\mathbf{W}^{(t)^{T}}\mathbf{W}^{(t)}$ ~\cite{boyd2006randomized}.

\begin{assumption}
[Bounded Heterogeneity]\label{assump:heter} There exists a constant $\tau_i >0$ for any node $i \in \mathcal{C}$, for any $\xi_i \in D_i$ and $\forall j\in \mathcal{C}, j\neq i ;\xi_j \in D_j$, such that
$ \mathbb{E} \left\| \nabla F_i(\mathbf{x}_i, \xi_i) - \frac{1}{N} \sum_j \nabla F_j(\mathbf{x}_j, \xi_j) \right\|^2 \leq \tau_i$.    
\end{assumption}

Assumption \ref{assump:heter} measures the heterogeneity and dissimilarity across the clients and their local objectives raised from the difference in local datasets.

\begin{assumption}
[Convexity]\label{assump:convx} 
Each local objective function $f_i: \mathbb{R}^d \rightarrow \mathbb{R}$ is convex. 
Formally, for all $\mathbf{x}, \mathbf{y} \in \mathbb{R}^{dim}$ and for all $i \in \{1, \ldots, N\}$,
\[
F_i(\mathbf{y}, \xi) \ge F_i(\mathbf{x},\xi) + \langle \nabla F_i(\mathbf{x}, \xi), \mathbf{y} - \mathbf{x} \rangle.
\]
\[
f_i(\mathbf{y}) \ge f_i(\mathbf{x}) + \langle \nabla f_i(\mathbf{x}), \mathbf{y}-\mathbf{x} \rangle.
\]
This implies that the global objective function $f(\mathbf{x}) = \frac{1}{N} \sum_{i=1}^{N} f_i(\mathbf{x})$ is also convex.   
\end{assumption}

\begin{property}
[B-Strong Connectivity]\label{prop:bstrong} 
A graph $\mathcal{G}_g = (\mathcal{V}, E)$ is said to be \emph{strongly connected} if, for every pair of nodes $(i, j) \in \mathcal{V}$, there exists a path from $i$ to $j$. Formally, a graph is B-strongly connected if there exists an integer $B>0$ such that the graph with the edge set $E_B = \bigcup_{t=kB}^{(k+1)B-1} E(t)$ is strongly connected for all $k \in \mathbb{N}$.   
\end{property}

Property \ref{prop:bstrong} is weaker than requiring the graph to be strongly connected at every round. It indicates that the graph accumulates the necessary edge set over time, allowing these edges to appear in an arbitrary order, as long as the overall graph becomes strongly connected after a certain number of rounds denoted by $B$~\cite{nedic2014distributed}. 
% \vspace{-3mm}
% \subsection{}

\vspace{-4mm}

\subsection{Mobility and B-Strong Connectivity}

\noindent In this section, we formally illustrate how mobility contributes to this property and further utilize it to prove the convergence in DFL. Although this property is satisfied under mobility, it does not hold for the static baseline where there is no mobile client.

\subsubsection{Static}
When all clients in the network maintain fixed locations throughout the entire training procedure, the graph retains a fixed set of edges, resulting in a constant mixing matrix $\mathbf{W}$. Formally:

\vspace{-6mm}

\begin{align}
\forall \ t\in[T];\ E(t)=E  \Rightarrow \mathbf{W}^{(t)}=\mathbf{W}^{(0)}=\mathbf{W}.
\end{align}

\vspace{-3mm}

In this DFL setting, the entries of the mixing matrices are determined based on each client's neighborhood, limited to a finite communication radius $R_c$. In sparse networks, e.g. when $R_c$ is small, there may exist isolated nodes and the graph is disconnected. As the graph and the mixing matrix remain fixed throughout the entire period, this sparsity persists, and it is implied that B-Strong Connectivity may not be satisfied as a sparse graph is maintained with the same set of edges during the entire training procedure.

\subsubsection{Mobile Network}
In such networks, mobile clients relocate to new locations during the training period. When there exists at least one mobile client in the network, the graph and mixing matrix will change over time. Considering \textbf{Random Movement} as the baseline, mobile clients choose each location in the network at least once after a certain number of rounds, since their trajectory is controlled by a uniform distribution across the network locations. This observation shows that all static clients establish a connection with the mobile client during the training period, ensuring that the graph is connected. 
% This insight can further help us design mobility strategies to develop more efficient DFL frameworks by exploiting additional information, favoring B-Strong Connectivity property and mitigating data heterogeneity.

\begin{figure}[t]
    \centering
    \includegraphics[width=0.8\linewidth]{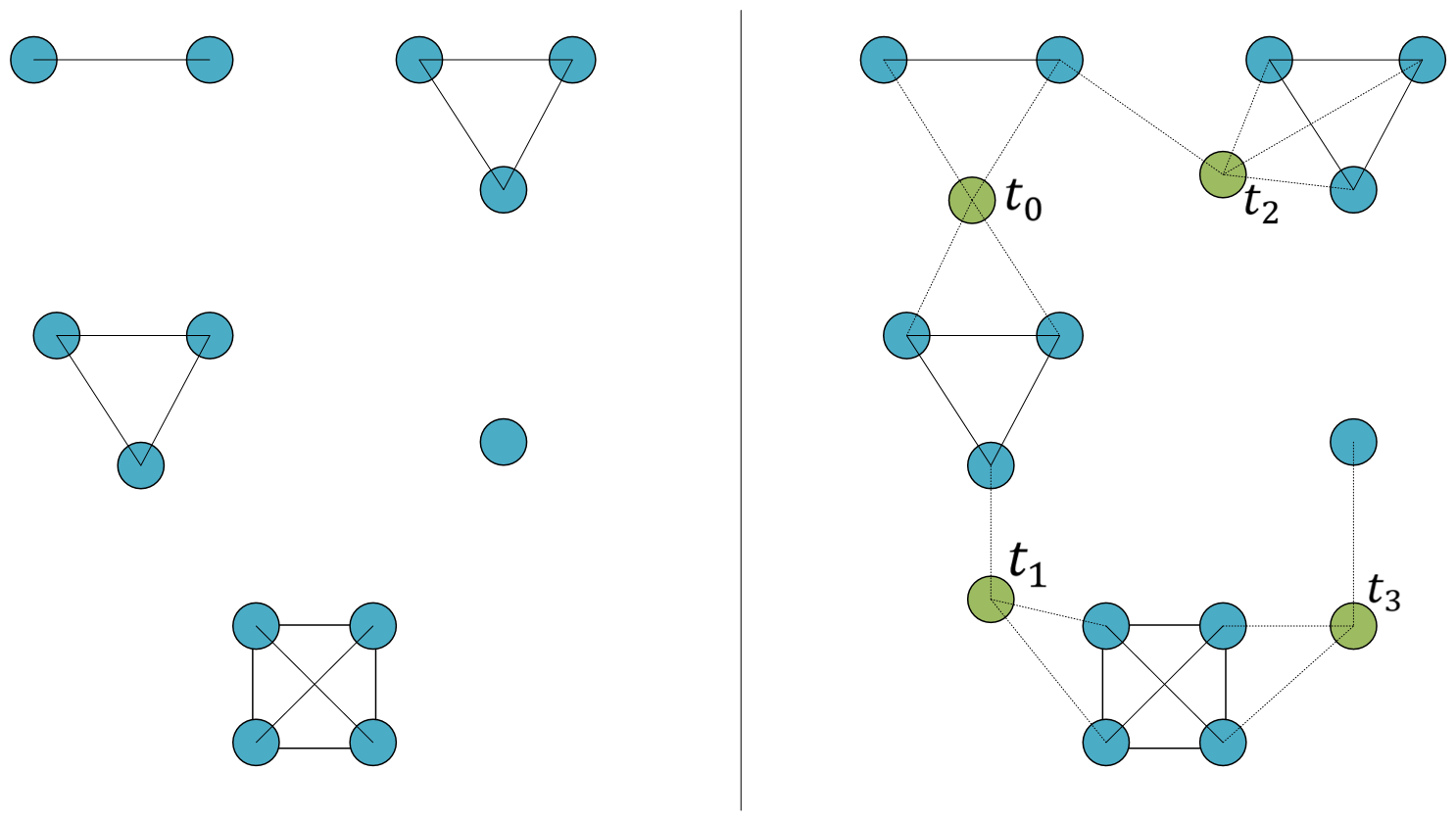}
    \caption{Graphs representing the connectivity of clients in a network with a limited communication radius respectively in a static and mobile network}
    \label{fig:graph}
\end{figure}

Fig.~\ref{fig:graph} represents the connectivity graph of multiple clients in a network operating in DFL system with the blue nodes denoting static clients and the green node denoting a mobile client. In the static scenario, the graph remains the same throughout the entire procedure, and the system suffers from sparsity. Despite the default sparsity in the network due to finite $R_c$, mobile nodes can establish connection with different clients each round. As illustrated in Fig. \ref{fig:graph}, the green node relocates to different locations in multiple rounds, establishing connection with the clients nearby. This illustration shows that mobile clients can ensure a uniform strong connection after a certain number of rounds depending on the mobility strategy they operate on. This mobility strategy can be further improved by reducing the complexity of trajectory decision and considering heterogeneity mitigation.

Motivated by this insight, we provide a representation for the matrix of concatenated models at round $t$, i.e. $\mathbf{X}^{(t)}$, based on the matrix of concatenated models at previous $B$ time steps and all the concatenated gradients in between from $t-B$ to $t$ in Lemma \ref{lem:stack}.

\vspace{-4mm}

\subsection{Convergence Analysis}
In this section, we establish the convergence bound of DFL under mobility. We start with deriving the needed Lemmas and Corollary, and then proceed with deriving the convergence of DFL system in Theorem \ref{th:theorem}.

\begin{lemma}
[Stacked models representation]\label{lem:stack} 
Let $\mathbf{X}^{(t)} = \big[ \mathbf{x}_1^{(t)},..., \mathbf{x}_N^{(t)} \big] \in \mathbb{R}^{dim \times N}$, $\mathbf{W}^{(t)}$ denote the doubly stochastic mixing matrix in round $t$, and $\mathbf{G}^{(t)} = [\mathbf{g}_1^{(t)},..., \mathbf{g}_N^{(t)}]$ where $\mathbf{g}_i^{(t)}=\nabla F_i (\mathbf{x}_i^{(t)},\xi_i^{(t)})$ denotes the mini-batch gradient of client $i$. Then, 
% \vspace{-5mm}
\begin{equation}\label{eq:stacked_model}
\mathbf{X}^{(t)} = \mathbf{X}^{(t-B)} \cdot \prod_{k=t-B}^{t-1} \mathbf{W}^{(k)} - \eta \sum_{k=t-B}^{t-1} \left[ \mathbf{G}^{(k)} \prod_{j=k}^{t-1} \mathbf{W}^{(j)} \right].
\end{equation}    
\end{lemma}

\vspace{-3mm}

We refer the proof of this lemma to the Appendix. Moreover, we further illustrate how close a B-strongly connected sequence of graphs can get to a graph with fully-connected topology, where all the nodes are connected, in Corollary \ref{th:corol2}. Lastly, we present the convergence bound of DFL systems under mobility in the network in Theorem \ref{th:theorem}. The proof of this corollary and theorem are referred to the Appendix.

\begin{corollary}
[Mixing Matrices Product]\label{th:corol2} 
Let $\bm \psi^{(B)}=\prod_{t=kB}^{(k+1)B-1}\mathbf{W}^{(t)}$ denote the product of mixing matrices at $B$ consecutive rounds, where $\mathbf{W}^{(t)}$ denotes the mixing matrix for a B-strongly connected sequence of graphs:

\vspace{-5mm}

\begin{align} \label{eq:psi_bound}
    \| \bm \psi^{(B)} - \frac{\mathbf{11}^\top}{N} \|_F^2 \leq C^2N^2\lambda^{2B},
\end{align}

\vspace{-2mm}

\noindent where we can always choose $\lambda=(1-\frac{1}{N^{NB}})^{\frac{1}{B}}$ and $C=4$\cite{nedic2014distributed}.    
\end{corollary}

Corollary \ref{th:corol2} measures how close the product of mixing matrices over $B$ consecutive rounds for a B-Strongly Connected sequence of graph can be to a fully connected graph.

\begin{theorem}
[Optimization error of global objective function]\label{th:theorem} 
Let $\bar{\mathbf{x}}^{(t)}=\frac{1}{N}\sum_{i\in \mathcal{C}}\mathbf{x}_i^{(t)}$ denote the average of all models, $\mathbf{x}^*$ denote the global minimizer and $f^*=f(\mathbf{x}^*)$, $\eta_t \leq \eta \leq \frac{1}{2L}$, $\bar{\sigma}^2 = \frac{1}{N}\sum_{i\in \mathcal{C}}\sigma_i^{2}$, and $\hat{\tau}=\sum_{i\in \mathcal{C}}\tau_i$. Under Assumptions \ref{assump:smooth}-\ref{assump:convx}, the optimization error for the DFL system with mobility is bounded as follows:

\vspace{-5mm}

\begin{align}\label{eq:main_bound}
&\frac{1}{T+1}\sum_{t=0}^T \mathbb{E}\big\{ f(\bar{\mathbf{x}}^{(t)}) - f^* \big\}
\leq \frac{1}{2\eta(1-2L\eta)}\\ & \Bigg(  \nonumber
\frac{\|\bar{\mathbf{x}}^{(0)} - \mathbf{x}^*\|^2}{T+1}
+ \frac{\eta^2 \bar{\sigma}^2}{N}   \nonumber
+ \frac{L\eta^3}{N}(2L\eta+1)A_0         \nonumber
\Bigg),
\end{align}

where

\vspace{-6mm}

\begin{align}
A_0 = \left(1 + \frac{2}{p}\right)\left(\frac{2}{p}\right)\hat{\tau}\big(\frac{(1-p)}{p} + C^2 N^2 \lambda^{2B}\big) .
\end{align}    
\end{theorem}

\noindent\textbf{Discussion.}
Lemma \ref{lem:stack} and Corollary \ref{th:corol2} illustrate how the movement of mobile clients dynamically influences network topology and facilitates information propagation. As expressed in Eq.~(\ref{eq:stacked_model}), the model at each round depends on the models from the preceding $B$ steps and the gradients computed within that interval. This relation indicates that time-varying topologies under mobility enhance information propagation across the network. For instance, the product of mixing matrices over time, allows client $i$ to use gradients from client $j$ even without a direct communication link. Corollary \ref{th:corol2} also captures the deviation between the accumulated graphs over time and a fully connected graph, showing that mobility can effectively compensate for sparsity.  

The term $\hat{\tau}$ in Theorem~\ref{th:theorem} reflects the direct influence of data heterogeneity on convergence. As data distributions become less heterogeneous, the bound tightens and convergence improves. While heterogeneity can degrade performance, its effect is coupled with graph connectivity, which can be reinforced through an appropriate mobility strategy that ensures uniform communication among clients. If mobile clients periodically establish connection with all static clients, i.e., entering their neighborhoods defined as  
$\mathcal{N}_i^{(t)} = \{j \mid \| L_i^{(t)} - L_j^{(t)}\|_2 \leq R_c, \forall j \in \mathcal{C}\setminus\{i\}\}$,  
the resulting sequence of graphs satisfies the \textbf{B-Strong Connectivity} property, guaranteeing convergence. Faster satisfaction of this condition yields smaller $B$, which reduces $\lambda = (1 - \frac{1}{N^{NB}})^{\frac{1}{B}}$, improving both convergence speed and accuracy while resolving the effects of data heterogeneity.

\vspace{-5mm}

\subsection{Methodology intuition and motivation} 
% \noindent The representation in Eq.~(\ref{eq:stacked_model}) illustrates the mechanism of information transfer within the network under mobility. To further enhance this process, we intuitively design the trajectory planning of mobile clients around two main objectives: (1) establishing connections with a greater number of static clients simultaneously, and (2) compensating for the lack of data among static clients across different classes. This representation indicates that a client can update its model using the gradient information of another client that possesses samples from classes absent in its own dataset or with a different distribution, even without a direct communication link. This gradient exchange mechanism mitigates training bias and consequently improves the overall performance of the DFL system.
\noindent The representation in Eq.~\eqref{eq:stacked_model} illustrates the mechanism of information propagation in DFL under mobility: a client can effectively incorporate gradient information of other clients in the absence of direct communication link. This observation suggests that the gradient exchange process can be facilitated through trajectory design. As the performance degradation in sparse DFL is largely attributed to data heterogeneity, mobility can be leveraged to transfer distinct information across regions with different class compositions. Class-wise distribution discrepancy between regions provides an effective measurement to determine how relocation of mobile clients can boost information transfer.

The convergence bound in Theorem~\ref{th:theorem} highlights the joint impact of data heterogeneity and network connectivity respectively associated with the terms $\hat{\tau}$ and $B$. This suggests that to achieve a smaller error bound, mobile node trajectory should be designed around two main objectives: (i) combating heterogeneity by guiding mobile clients towards regions with a larger class-wise distribution discrepancy compared to current location, and (ii) improving connectivity by selecting regions that allow simultaneous interaction with a larger number of static clients to achieve a smaller $B$. 

Motivated by these findings, we propose a mobility-assisted DFL framework in section \ref{sec:framework}, which enables mobile clients to move strategically across the network. The proposed methodology aims to achieve near-uniform connectivity and enhance information exchange among heterogeneous clients. Building upon this theoretical foundation, Section \ref{sec:framework} introduces our \textit{Distribution-Aware Mobility} strategies, which leverage knowledge of local data distributions and spatial positions to determine trajectories that maximize information transfer and mitigate non-IID data effects. 

% Explicitly, mobile nodes determine their path to maximize the dissemination of new information. They move toward regions where nearby static clients have not seen the data classes the mobile client just learned. To further enhance movement efficiency, we introduce a refined version of this strategy that enables the mobile client to move toward locations where it can establish connections with multiple static clients simultaneously.

% built on the insight discussed in section~\ref{sec:theory}

\vspace{-2mm}

\section{Distribution Aware Mobility-Assisted DFL}\label{sec:framework}
\noindent In this section, we propose our novel mobility-assisted DFL framework to further improve the information flow in the network. 
% It is common in the FL literature to treat client model gradients as representative information of their local data. For instance, the authors of~\cite{ye2023feddisco} employ class-wise information in aggregation while relying on gradients for theoretical modeling. 
% Following a similar approach, we design our methodology based on local data information to enable mobile clients to facilitate gradient exchange.

\vspace{-4mm}

\subsection{Distribution Distance}
\noindent In FL, it is commonly assumed that the central server is aware of the mobile clients speed and locations \cite{yu2020mobility}. Additionally, authors in \cite{ye2023feddisco} assume that the server knows each client's category-wise data distribution. Following a similar approach in this work, we assume that mobile clients collect the information of static clients' locations and category-wise data distribution through a warm-up phase.
In practice, the required metadata can be transmitted through a lightweight control-plane mechanism. Specifically, each static client reports location and class-wise distribution (label histogram) once at initialization to a backend entity, which broadcasts this aggregated metadata to mobile clients without participating in model training or aggregation. This information exchange is lightweight (e.g. a few bytes per clients), preserves data privacy since no raw samples or gradients are shared, and is performed only once prior to training. 
Assuming 10 classes, this metadata consisting of the client location (two coordinates, 64 bits) and the class-wise label histogram (10 classes, 320 bits) amounts to a total of 384 bits (48 bytes) per client.

These assumptions allow mobile clients to calculate the aggregated data distribution of static clients within the communication range of different locations across the network. 
Given the category-wise distribution of static clients and their locations, a mobile client $i_m \in \mathcal{C}_m$ can determine the aggregated data distribution at any location in $\mathcal{G}$ before starting training. 
The category-wise distribution at a location $L \in \mathcal{G}$, if the mobile client moves there, is given by: \vspace{-1mm}
\begin{equation} \label{eq:data_dist}
    D_{i_m,L}(l) = \frac{\sum_{j \in \mathcal{S}_L \cup \{i_m\}} |\{(\mathbf{s},y) \in \mathcal{D}_j : y=l\}|}{\sum_{j \in \mathcal{S}_L \cup \{i_m\}} |\mathcal{D}_j|}, \quad l \in [Y],
\end{equation}
where $\mathcal{S}_L := \{j \mid \|L_j^{(0)} - L \|_2 \leq R_c, j \in \mathcal{C}_s\}$ denotes the set of fixed clients within the coverage area of the mobile client $i_m$ at location $L$ and $[Y]$ denotes the set of classification labels.

To measure heterogeneity between neighborhoods, we use this category-wise distribution and define the \textit{distribution distance} from the client's current location $L'$ (considered as the reference point) as:

\vspace{-6mm}

\begin{equation} \label{eq:dist_diff}
    d(L,L') = \Big(\sum_{l \in [Y]} \big(D_{i_m,L}(l)-D_{i_m,L'}(l)\big)^2\Big)^{1/2}.
\end{equation}

\vspace{-10mm}

\subsection{Distribution-Aware Mobility Strategies}
\noindent Mobile clients are allowed to use the information in Eq.~(\ref{eq:dist_diff}) to design their trajectory considering the heterogeneity difference between the neighborhoods. The mobility strategies discussed in this section effectively mitigate the impact of data heterogeneity by utilizing the mobile clients to move from one location to another with the highest possible distribution difference, thus transferring completely distinct information to the latter neighborhoods with a higher probability. 

\subsubsection{Distribution Aware Movement (DAM)}
Following the insight on maximizing information transfer by moving from one location to a new one each round, we formally propose a mobility strategy that allows mobile clients to take advantage of category-wise distribution knowledge in addition to clients location knowledge to plan their waypoint throughout the training.
The key idea of DAM is to guide the mobile clients to move toward neighborhoods with the most different aggregated data distribution from their current locations, enabling them to act as relays or bridges to transfer information and reduce the distribution imbalance across the network.

At round $t$, a mobile client at $L_{i_m}^{(t)}$ assigns probability
\vspace{-2mm}
\begin{equation} \label{eq:prob_dist}
    p(L|L_{i_m}^{(t)}) = \frac{d(L,L_{i_m}^{(t)})}{\sum_{L' \in \mathcal{G}} d(L',L_{i_m}^{(t)})}, \quad L \in \mathcal{G},
\end{equation}
to move toward $L$, thereby favoring locations with more distinct distributions. The desired location is sampled from the distribution acquired in Eq.~(\ref{eq:prob_dist}), i.e.  $L_{i_m}^{(des,t)}\sim p(L|L_{i_m}^{(t)})$.
If the chosen $L_{i_m}^{(des,t)}$ lies beyond $R_m$, the client moves to the nearest feasible location along that direction.
Due to the mobility constraint $R_m$, the actual update is
\begin{equation}
     L_{i_m}^{(t+1)} = \begin{cases}
         L_{i_m}^{(des., t)}; \quad \text{if } \|L_{i_m}^{(des., t)}  - L_{i_m}^{(t)}\|_2 \leq R_m \\
        \argmin_{L \in \Gamma} \|L-L_{i_m}^{(des., t)} \|_2; \quad \text{otherwise},
     \end{cases}
     \label{eq:obtained_loc}
\end{equation}
where $\Gamma=\{L \in \mathcal{G}\mid \|L-L_{i_m}^{(t)}\|_2 \leq R_m\}$. If the mobile client does not reach the destination, within one round, it remains active until achieved, after which a new target is drawn. This encourages exploration of regions with distinct distributions, thereby enhancing information mixing across heterogeneous neighborhoods while respecting mobility limits. 

\subsubsection{Distribution-Aware Cluster Center Movement (DCM)}

DAM considers all grid points, which can be inefficient in large networks. DCM reduces complexity by restricting movement to cluster centers of static clients, providing a smaller but more informative set of destinations.

Clusters are formed iteratively: the first center is the location covering the largest number of static clients within $R_c$; subsequent centers are chosen from uncovered clients, prioritizing coverage of most static clients. This yields a compact set of representative destinations $\mathcal{L}_c$. Mobile clients then apply the same distribution-aware rule as in DAM, but only over $\mathcal{L}_c$: \vspace{-2mm}
\begin{equation} \label{eq:DCM_prob_dist}
    p(L_c|L_{i_m}^{(t)}) = \frac{d(L_c,L_{i_m}^{(t)})}{\sum_{L_c' \in \mathcal{L}_c} d(L_c',L_{i_m}^{(t)})}, \quad L_c \in \mathcal{L}_c.
\end{equation}

By focusing on cluster centers, DCM accelerates convergence while retaining the benefits of distribution-aware movement. 
The clustering process is summarized in Algorithm~\ref{alg:clustering}.

\noindent\textbf{Complexity Scaling.}
In DAM and the COM baseline, a mobile client respectively evaluates the distribution distance in Eq. \eqref{eq:dist_diff} and node degree difference over all candidate locations in the grid, i.e., $|\mathcal{G}|=G^2$ candidates, whereas DCM restricts the candidate set to the cluster centers $\mathcal{L}_c$ generated by Algorithm~\ref{alg:clustering}, i.e., $|\mathcal{L}_c|$ candidates. Therefore, the search complexity is reduced from $\mathcal{O}(G^2)$ to $\mathcal{O}(|\mathcal{L}_c|)$, yielding a reduction factor of $\frac{G^2}{|\mathcal{L}_c|}$. Using a simple approximation, the number of clusters can be estimated as the number of circular areas with radius $R_c$ required to cover the network, i.e., $|\mathcal{L}_c| \leq \frac{G^2}{\pi R_c^2}$. Accordingly, the reduction scales as $\frac{G^2}{|\mathcal{L}_c|}\geq \pi R_c^2$. This estimate assumes the worst case where the entire network is clustered; in practice, the reduction is typically larger.

\begin{algorithm} \small
	 \KwIn{Communication radius $R_c$; Grid $\mathcal{G}$; Static clients $\mathcal{C}_s$; Location of static clients $\mathcal{L}_s$.}
	 \KwOut{Cluster centers $\mathcal{L}_c$.} 

    \tcp{Time Complexity: $\mathcal{O}(|\mathcal{G}|^2\cdot |\mathcal{C}_s|^2$)}
     
	 $\mathcal{L}_c \gets \{\}$; $\mathcal{C}_s^r \gets \mathcal{C}_s$; $n=1$\\
     \While{$|\mathcal{C}_s^r| > 0$}{
        $L_{c_n} \sim \argmax_{L \in \mathcal{G}}|\{j \in \mathcal{C}_s^r \mid \| L - L_j^{(0)} \|_2 \leq R_c\}|$ \\
        $\mathcal{C}_s^r = \mathcal{C}_s^r \setminus \{j \in \mathcal{C}_s \mid \| L_{c_n}-L_j^{(0)} \|_2 \leq R_c\}$ \\
        $\mathcal{L}_c = \mathcal{L}_c \cup \{L_{c_n}\}$; $n \gets n+1$
     }
    \caption{Clustering}
    \label{alg:clustering}
\end{algorithm}

% \vspace{-2mm}

\begin{algorithm}[t]\small
\caption{DistMove}
\label{alg:dam}
\KwIn{Mobile client ID $i_m$; Mobility Constraint $R_m$; Communication Radius $R_c$; Current Location $L_{i_m}^{(t)}$; Movement Mode ($MV$); Grid Locations $\mathcal{G}$; Cluster Centers $\mathcal{L}_c$; Static clients and their locations $\mathcal{C}_s$, $\mathcal{L}_s$}
\KwOut{Next location $L_{i_m}^{(t+1)}$}
\tcp{Time Complexity: \\ $\mathcal{O}(|\mathcal{G}|^2)$ for DAM, $\mathcal{O}(|\mathcal{L}_c|)$ for DCM}

\If{$MV = \text{DAM}$}{
    $\mathcal{S} \gets \mathcal{G}$\;
}
\If{$MV = \text{DCM}$}{
    $\mathcal{S} \gets \mathcal{L}_c$\;
}

\If{$flag = 0$ \textbf{ or } $t = 0$}{
    \For{$L \in \mathcal{S}$}{
        $\mathcal{S}_L \gets \{ j \mid \|L_j^{(0)} - L\|_2 \leq R_c,\ j \in \mathcal{C}_s \}$\
        \For{$l \in [Y]$}{
            $D_{i_m,L}(l) = 
            \dfrac{
                \sum_{j \in \mathcal{S}_L \cup \{i_m\}}
                |\{(\mathbf{s},y) \in \mathcal{D}_j : y = l\}|
            }{
                \sum_{j \in \mathcal{S}_L \cup \{i_m\}} |\mathcal{D}_j|
            }$\
        }
        $d(L, L_{i_m}^{(t)}) =
        \big(\sum_{l \in [Y]} \big( D_{i_m,L}(l) - D_{i_m,L_{i_m}^{(t)}}(l) \big)^2\big)^{1/2} $\
    }

    \For{$L \in \mathcal{S}$}{
        $p(L \mid L_{i_m}^{(t)}) =
        \dfrac{
            d(L, L_{i_m}^{(t)})
        }{
            \sum_{L' \in \mathcal{G}} d(L', L_{i_m}^{(t)})
        }$\
    }

    $L_{i_m}^{(\text{des.},t)} \sim p(L \mid L_{i_m}^{(t)})$\;
}

\If{$\|L_{i_m}^{(\text{des.},t)} - L_{i_m}^{(t)}\|_2 \leq R_m$}{
    $L_{i_m}^{(t+1)} \gets L_{i_m}^{(\text{des.},t)}$\;
}
\Else{
    $\Gamma=\{L \in \mathcal{G}\mid \|L-L_{i_m}^{(t)}\|_2 \leq R_m\}$
    $L_{i_m}^{(t+1)} \gets 
    \arg\min_{L \in \Gamma} \|L - L_{i_m}^{(\text{des.},t)}\|_2$\
}

\If{$L_{i_m}^{(t+1)} = L_{i_m}^{(\text{des.},t)}$}{
    $flag \gets 0$\
}
\If{$L_{i_m}^{(t+1)} \neq L_{i_m}^{(\text{des.},t)}$}{
    $flag \gets 1$\;
    $L_{i_m}^{(\text{des.},t+1)} \gets L_{i_m}^{(\text{des.},t)}$\
}

\end{algorithm}

\begin{algorithm} \small
    \KwIn{Mobile Client ID $i_m$; Current Location $L_{i_m}^{(t)}$; Mobility Constraint $R_m$; Communication Radius $R_c$; Static Clients and their locations $\mathcal{C}_s,\mathcal{L}_s$; Grid Locations $\mathcal{G}$; Cluster Centers $\mathcal{L}_c$; Movement Mode($MV$)}
    \KwOut{Next location $L_{i_m}^{(t+1)}$}

    \tcp{Time Complexity: \\
    $\mathcal{O}(|\mathcal{G}|^2)$ for DAM, COM\\
    $\mathcal{O}(|\mathcal{L}_c|)$ for DCM
    }

    \If{$MV=$ Random Movement}{
        $L_{i_m}^{(t+1)} \sim \text{Uniform}\Big(\{L \mid \|L-L_{i_m}^{(t)}\|_2 \leq R_m, L \in \mathcal{G} \}\Big).$ \\
    }
    \If{$MV= DAM $ $||$ $ MV$= DCM}{
        $L_{i_m}^{(t+1)}=DistMove(i_m,R_m,L_{i_m}^{(t)},MV,\mathcal{G},\mathcal{L}_c,\mathcal{C}_s,\mathcal{L}_s)$
    }

    \If{$MV = COM$}{
    \If{$flag = 0$}{
    {
    $deg^{(t)}_{i_m} \leftarrow |\mathcal{N}^{(t)}_{i_m} \cap C_s|$ \\
    }
    \For{$L \in G$}{
        $S_L \leftarrow \{ j \mid \|L^{(0)}_j - L\|_2 \leq R_c, j \in C_s \}$ \ \\
        $deg_L \leftarrow |S_L|$ \\ 
        $\Delta deg_{i_m}^{(L,t)} \leftarrow deg_L - deg^{(t)}_{i_m}$ \\
    }
    
    {
        $p(L \mid L^{(t)}_{i_m}) = \frac{e^{\Delta deg_{i_m}^{(L,t)}}}{\sum_{L' \in G} e^{\Delta deg_{i_m}^{(L,t)}}}$
    }
    
    $L^{(des.,t)}_{i_m} \sim p(L \mid L^{(t)}_{i_m})$ \
    }

    \If{$\|L^{(des.,t)}_{i_m} - L^{(t)}_{i_m}\|_2 \le R_m$}{
        $L^{(t+1)}_{i_m} \leftarrow L^{(des.,t)}_{i_m}$
    }
    \Else{
        $\Gamma = \{ L \in G \mid \|L - L^{(t)}_{i_m}\|_2 \le R_m \}$ \
        $L^{(t+1)}_{i_m} \leftarrow \arg\min_{L \in \Gamma} \|L - L^{(des.,t)}_{i_m}\|_2$
    }
    \If{$L_{i_m}^{(t+1)} = L_{i_m}^{(\text{des.},t)}$}{
    $flag \gets 0$\
    }
    \If{$L_{i_m}^{(t+1)} \neq L_{i_m}^{(\text{des.},t)}$}{
    $flag \gets 1$\;
    $L_{i_m}^{(\text{des.},t+1)} \gets L_{i_m}^{(\text{des.},t)}$\
    }

}
    \caption{Client Mobility}
    \label{alg:mobility}
\end{algorithm}

\noindent \textbf{Design Discussion.}  
% For probability assignment (Eqs.~(\ref{eq:prob_dist}) and~(\ref{eq:DCM_prob_dist})), we adopt simple normalization, which empirically outperformed softmax in our experiments.
We employ the $\ell_2$ distance to measure distribution discrepancy since it provides a symmetric and bounded metric over class-wise histograms, while avoiding singularities that may arise in divergence-based measures such as KL when class probabilities are zero. For probability assignment, we adopt simple linear normalization rather than softmax to preserve the proportional relationship between candidate discrepancies without introducing additional nonlinear scaling or temperature hyperparameters. 
% To empirically validate the effectiveness of our design choices, we report the final accuracy under an experimental setup in table \ref{tab:soft_norm}.

% \begin{table}[h]
% \centering
% \caption{Comparing normalization and softmax for MNIST experiment with 20 clients in a network with 3 mobile clients. \vspace{-2mm}}
% \begin{tabular}{|c|c|c|}
% \hline
% \textbf{Movement Pattern} & Softmax & Normalization \\
% \hline
% DCM & 77.80\% & 80.83\% \\
% \hline
% DAM & 77.24\% & 79.85\% \\
% \hline
% \end{tabular}
% \label{tab:soft_norm} \vspace{-2mm}
% \end{table}

% This design maintains stable exploration toward regions with larger distribution differences while keeping the mobility control lightweight and robust.

Algorithms~\ref{alg:dam} and~\ref{alg:mobility} illustrate how the movement of mobile clients is controlled according to the pre-defined mobility pattern. These mobility patterns consist of the baseline model, \textbf{Random Movement} and \textbf{COM}, and our proposed mobility strategies, \textbf{DCM} and \textbf{DAM}. Although we consider a grid-based mobility model for clarity, the proposed framework is not restricted to grid networks and can be naturally extended to graph-based mobility settings (e.g., road networks) by redefining the feasible location set and using shortest-path distance over the mobility graph. In such cases, the selected destination is mapped to a reachable node in the mobility graph and the mobile client follows a shortest-path routing policy, while the connectivity-based convergence analysis remains the same.

% \noindent\textbf{Discussion.} COM baseline and DAM mobility strategies enhance the information flow by respectively considering graph connectivity and class-wise data distribution. DCM aims to target both connectivity and data distribution with the benefit of complexity reduction. Both DAM and DCM extend the role of mobile clients beyond random relays. DAM flexibly explores all feasible locations, while DCM improves efficiency by limiting movements to representative centers. In both cases, mobility strategically connects sparse or heterogeneous regions, enhancing information dissemination and accelerating convergence.

% %%% Revised and repolished till here %%%

% \noindent\textbf{Impact on Connectivity Parameter $B$.}
% While Theorem \ref{th:theorem} shows that convergence depends on both the data heterogeneity term $\hat{\tau}$ and the connectivity parameter $B$, the proposed mobility strategies primarily improve information transfer considering the distribution discrepancy. However, DCM inherently balances both distribution discrepency and connectivity by directing mobile clients to cluster centers that simultaneously cover many static clients and exhibit large distribution differences, thus promoting faster information mixing. As empirically validated in Section~\ref{sec:exp}, DCM reduces not only the impact of data heterogeneity but also the effective $B$, confirming our theoretical intuitions.

\noindent\textbf{Discussion.}
The COM baseline and DAM enhance information flow by respectively prioritizing graph connectivity and class-wise distribution discrepancy, while DCM jointly considers both aspects with the additional benefit of reduced search complexity. DAM explores all feasible locations to transfer heterogeneous information, whereas DCM guides mobile clients toward representative cluster centers, simultaneously connecting to a large number of static clients and transferring distinct information across heterogeneous regions. In both cases, mobile nodes strategically connects sparse or heterogeneous regions, enhancing information dissemination and accelerating convergence. Although Theorem~\ref{th:theorem} indicates that convergence depends on both the heterogeneity term $\hat{\tau}$ and the connectivity parameter $B$ and DAM operates only based on distribution discrepancy, DCM inherently balances both factors. As empirically validated in Section~\ref{sec:exp}, DCM reduces not only the impact of data heterogeneity but also parameter $B$ effectively, confirming our theoretical intuitions.

\section{Experimental Results} \label{sec:exp}

\noindent In this section, we conduct experimental evaluations to assess the performance of DFL, validate the superiority of our proposed method, and further investigate the impact of various network parameters on overall performance.
Our experiments follow several directions, which we briefly introduce below before presenting the results.

% the effects of

\begin{itemize}
    \item We empirically demonstrate how mobility improves DFL performance.
    \item We first show that the proposed DCM and DAM strategies further enhance performance by mitigating data heterogeneity, and then validate the superiority of our method compared to the baseline under different network configurations.
    \item We conduct extensive simulations by varying key network parameters, i.e., $R_c$, $|\mathcal{C}_m|$, and $R_m$, and provide a comprehensive analysis of their impact to validate the theoretical intuitions discussed in Section~\ref{sec:theory}.
    \item We extend our simulations to larger networks and additional datasets to evaluate the robustness and generalization capability of the proposed framework.

    %Empirically demonstrating DCM decreases comm. bottleneck, and |Cm| deployment expenses
\end{itemize}

\subsection{Simulation Settings}
\noindent We conduct the experiments on the MNIST and CIFAR-10 classification tasks under a decentralized federated learning framework. Unless specified, we consider a network of size $18 \times18$ with $|\mathcal{C|}=20$ clients. Initial locations of clients are drawn from a uniform distribution across the locations in the network and a subset $\mathcal{C}_m \subset \mathcal{C}$ is assigned as mobile clients. To model data heterogeneity, training data samples are distributed across the clients following a Dirichlet distribution~\cite{lin2016dirichlet} controlled via parameter $\alpha$. Higher values of $\alpha$ indicate less heterogeneous scenarios, while lower values of $\alpha$ highlight higher heterogeneity. The test set of the corresponding dataset is used for evaluation across clients.

For MNIST, each client trains a simple CNN (two convolutional and two fully connected layers) using full-batch gradient descent with a learning rate of 0.3 for 1,000 rounds. For CIFAR-10, we adopt ResNet18~\cite{he2016deep} pretrained on ImageNet~\cite{deng2009imagenet}, with its first convolution layer modified (kernel size 3, stride 1), the MaxPooling layer removed, and trained with SGD (batch size 512, learning rate 0.1, momentum 0.9, weight decay $5\times10^{-4}$) for 2,000 rounds. Communication between clients is restricted to neighbors within radius $R_c$, and mobile clients are subject to a maximum displacement $R_m$ per round. Unless noted otherwise, $|\mathcal{C}_m|=3$, $R_c=3$, and $R_m=5$.

Performance is reported as the average test accuracy across clients, after 1,000 and 2,000, rounds, respectively for MNIST and CIFAR10 dataset aggregated over 6 Monte Carlo trials with the shaded regions in some figures showing the standard deviation around the mean accuracy across the trials. Reported accuracy in the figures for comprehensive analysis is the maximum accuracy acquired in that specific setting during the training round. Specific hyperparameters used in each experiment are reported in the titles above the corresponding figures for clarity.

\subsection{Results} \label{sec:results}

\subsubsection{Mobility Impact}
\begin{figure}[t]
    \begin{subfigure}{0.24\textwidth}
        \includegraphics[width=0.9\textwidth]{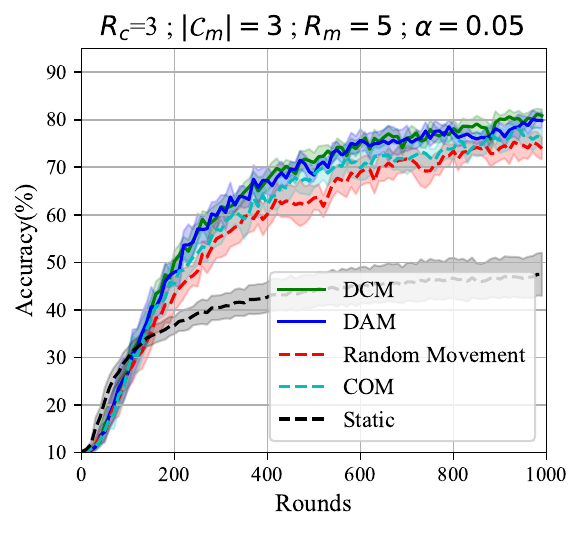}
        \caption{$\alpha=0.05$}
        \label{fig:acc_cons_mnist_alpha0.05}
    \end{subfigure}
    \begin{subfigure}{0.24\textwidth}
        \includegraphics[width=0.9\textwidth]{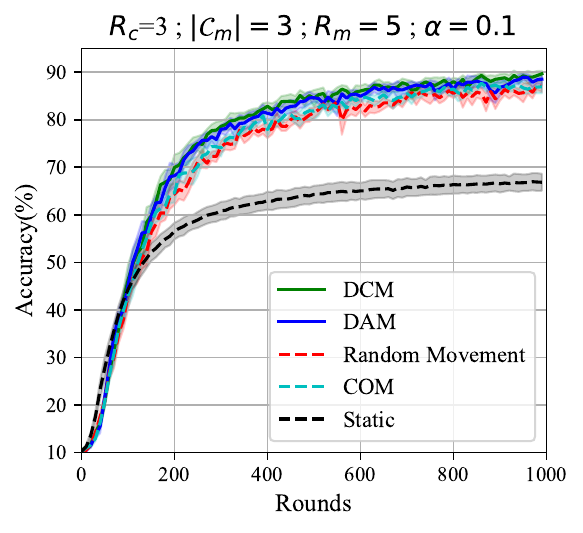}
        \caption{$\alpha=0.1$}
        \label{fig:acc_cons_mnist_alpha0.01}
    \end{subfigure}
    \caption{Test accuracy on MNIST dataset with 20 clients.}
    \label{fig:mnist_acc}
\end{figure}

Fig.~\ref{fig:mnist_acc} shows the test accuracy of DFL on MNIST with 3 mobile clients in a network under two heterogeneous data settings (i.e., $\alpha=0.05$ and $\alpha=0.1$). These results show that even random mobility significantly improves accuracy over the static baseline by allowing mobile clients to act as relays between disconnected regions, confirming that mobility compensates for sparse connectivity.

\subsubsection{Performance Improvement with DAM and DCM}
Observed in Fig.~\ref{fig:mnist_acc}, both distribution-aware strategies consistently outperform COM and random movement. By guiding mobile clients toward regions with distinct neighborhood distributions, DAM and DCM improve information mixing and reduce the impact of data heterogeneity. Under highly non-IID data ($\alpha=0.05$), DCM and DAM respectively achieve $8\%$ and $5\%$ higher accuracy than random mobility; under milder heterogeneity ($\alpha=0.1$), the gain remains around $4\%$. DCM provides the best overall accuracy, benefiting from its structured set of candidate destinations.

\begin{table}[h]
\centering
\caption{Final test accuracy (\%) on MNIST with 20 clients in a network with 3 mobile clients. \vspace{-2mm}}
\begin{tabular}{|c|c|c|}
\hline
\textbf{Movement Pattern} & $\alpha=0.05$ & $\alpha=0.1$ \\
\hline
DCM & 80.83\% & 89.65\% \\
\hline
DAM & 79.85\% & 88.51\% \\
\hline
COM & 76.05\% & 86.91 \% \\
\hline
Random Movement& 72.90\% & 86.90\% \\
\hline
Static & 47.50\% & 66.84\% \\
\hline
\end{tabular}
\label{tab:client_init} \vspace{-2mm}
\end{table}

\begin{figure*}[t]
    \begin{subfigure}{0.32\textwidth}
        \centering
        \includegraphics[width=0.7\linewidth]{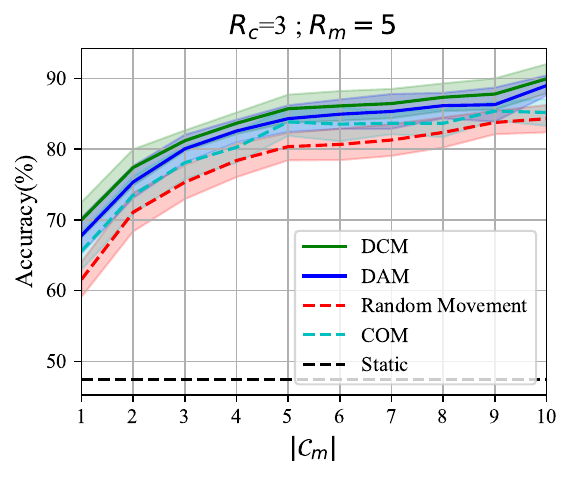}
        \caption{Accuracy vs. $|\mathcal{C}_m|$}
        \label{fig:mc}
    \end{subfigure} \hfill
    \begin{subfigure}{0.32\textwidth}
        \centering
        \includegraphics[width=0.7\linewidth]{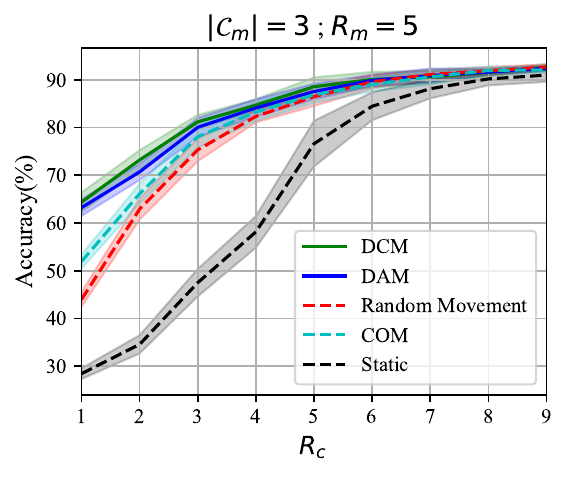}
        \caption{Accuracy vs. $R_c$}
        \label{fig:rc}
    \end{subfigure}
    \begin{subfigure}{0.32\textwidth}
        \centering
        \includegraphics[width=0.7\textwidth]{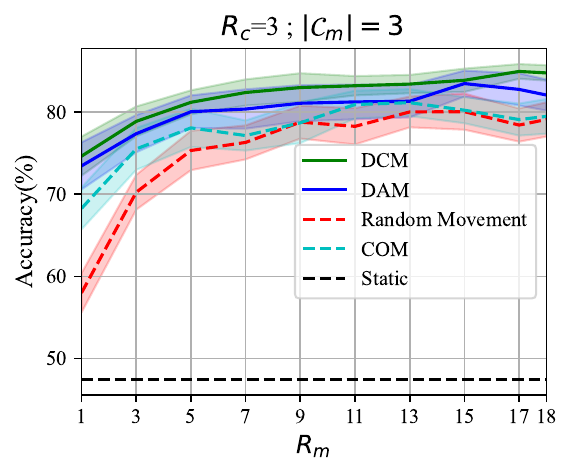}
        \caption{Accuracy vs. $R_m$}
        \label{fig:rm}
    \end{subfigure}
    \caption{Robustness of the method in any configuration of $\big(|\mathcal{C}_m|$, $R_c$ and $R_m\big)$ using MNIST dataset with 20 clients.} 
    \label{fig:Acc-param}
\end{figure*}

\subsubsection{Resolving Data Heterogeneity}
Table~\ref{tab:client_init} reports final accuracy for two heterogeneity levels ($\alpha=0.05$ vs.\ $\alpha=0.1$) under $R_c=3$, $R_m=5$, and $|\mathcal{C}_m|=3$. While static and random mobility baselines degrade sharply ($19.34\%$ and $14.00\%$ drops, respectively) when data becomes more heterogeneous, DAM and DCM limit the accuracy drop to under $10\%$. This confirms that distribution-aware mobility effectively mitigates the impact of non-IID data. COM baseline also limits the performance degradation under non-IID data to around $10\%$, suggesting the importance of connectivity enhancement in network.

\subsubsection{Other Network Configurations}
Fig.~\ref{fig:Acc-param} demonstrates the robustness of our proposed DFL framework under different network configurations, showing consistent advantage of DCM and DAM over the baselines (i.e., Random Movement and COM).

Fig.~\ref{fig:Acc-param}(\subref{fig:mc}) illustrates that increasing $|\mathcal{C}_m|$ improves accuracy, as more mobile clients serve as relays and disseminate information across the network. Both DAM and DCM achieve larger gains over random mobility and COM as $|\mathcal{C}_m|$ increases, highlighting the benefits of distribution-aware strategies in multi-mobile scenarios.
As the communication radius $R_c$ decreases, the gap between static and mobile cases widens (Fig.~\ref{fig:Acc-param}(\subref{fig:rc})), showing the advantage of DCM and DAM over random movement and the COM in sparse networks. In sparse topologies (e.g., $R_c=1$), DAM and DCM achieve up to $20\%$ higher accuracy than baselines by bridging disconnected regions more appropriately. As expected, larger $R_c$ values bring both baselines and the proposed strategies closer to this limit, with diminishing relative gains from mobility.
Fig.~\ref{fig:Acc-param}(\subref{fig:rm}) also examines the impact of $R_m$. The test accuracy improves with a larger $R_m$, since mobile clients can cover broader regions in one single step. Importantly, even with a smaller $R_m$, distribution-aware mobility provides a clear advantage over random mobility.

\noindent\textbf{Discussion.} Fig.~\ref{fig:Acc-param} empirically validates the theoretical insights from Section~\ref{sec:theory}. Faster satisfaction of strong connectivity (i.e., smaller $B$) leads to a lower error bound. Among all parameters, $R_c$ is the most critical, as it directly determines clients connectivity. Increasing $R_c$ naturally enhances connectivity and reduces $B$.  
The number of mobile clients $|\mathcal{C}_m|$ also significantly impacts performance, as mobile clients act as relays that bridge disconnected network regions. A larger $|\mathcal{C}_m|$ accelerates the achievement of strong connectivity, improving overall convergence.  
Similarly, the mobility constraint $R_m$ influences performance by limiting movement range. Higher $R_m$ values allow clients more freedom to reach target locations and establish wider connectivity.

Although these parameters directly affect performance, a trade-off exists between performance gains and practical limitations. Increasing $R_c$ introduces higher communication overhead, while larger values of $|\mathcal{C}_m|$ and $R_m$ incur higher deployment costs. This balance between efficiency and practicality is crucial for real-world deployment. Motivated by this trade-off, we proceed to the next experimental section.

\subsubsection{Fully-Connected Graph as upper bound}\label{sec:fc}
The experiments in this section are intended to empirically demonstrate the deployment efficiency of our proposed method compared to Random Movement and examine the trade off between network parameters and performance. Due to the large complexity of COM as a baseline and DAM, we limit these experiments to only DCM and Random Movement. We introduce a fully-connected scenario in which there is no communication constraint ($R_c=\infty$) and all clients can freely exchange their updates (fully-connected topology). Despite the rich connectivity among clients in the network and maximum accuracy, this topology dramatically increases the communication bottleneck similar to server-based FL. 

Considering this topology as an upper bound, we conduct extensive analysis to explore the trade offs between the parameters across the mobility strategies. To this end, we vary the parameters to determine the threshold value for $R_c$ and $|\mathcal{C}_m|$ required for the DFL system to achieve performance comparable to $90\%$ of a system operating under a fully-connected topology.

\begin{figure}[h]
    \begin{subfigure}{0.24\textwidth}
        \includegraphics[width=0.9\linewidth]{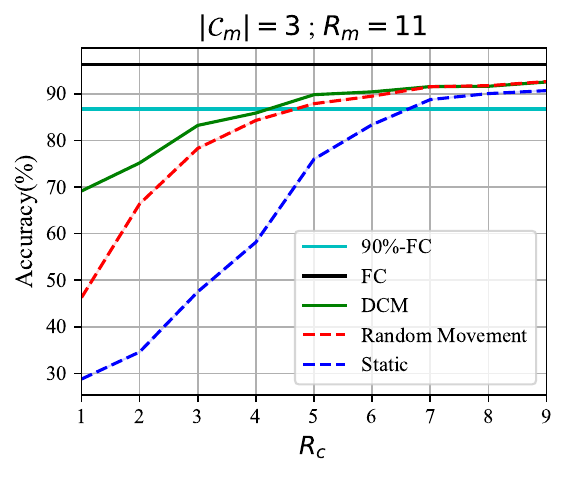}
        \caption{}
        \label{fig:FC_Rc1} 
    \end{subfigure}
    \hfill
    \begin{subfigure}{0.24\textwidth}
        \centering
        \includegraphics[width=0.9\linewidth]{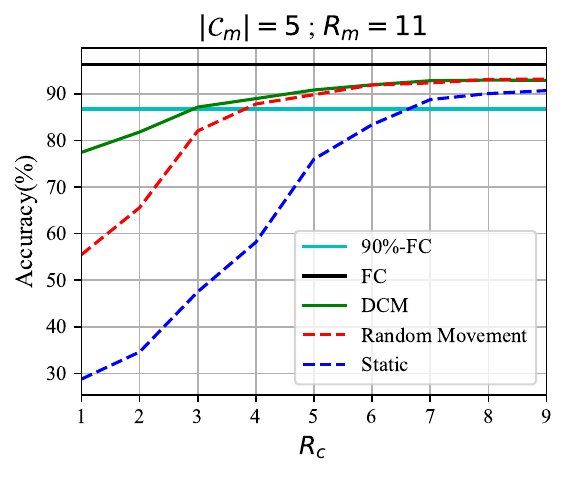}
        \caption{}
        \label{fig:FC_Rc2} 
    \end{subfigure}
    \caption{$R_c$ threshold in a network with 20 clients.}
    \label{fig:FC_Rc}
\end{figure}

Fig.~\ref{fig:FC_Rc} illustrates the performance curve with respect to $R_c$ with the upper bound discussed above. As shown in Fig.~\ref{fig:FC_Rc1}, DCM reaches the upper bound when $R_c=4$, whereas Random Movement requires $R_c=5$ under the same configuration with $|\mathcal{C}_m|=3$ mobile clients deployed. Similarly, in the same network, when $|\mathcal{C}_m|=5$, Random Movement requires $R_c=4$ to achieve $90\%$ of the fully connected performance, while DCM operates with $R_c=3$ and achieves higher accuracy. These findings empirically demonstrate that under similar configurations, DCM can effectively reduce the communication bottleneck in the network.

\begin{figure}[h]
    \begin{subfigure}{0.24\textwidth}
        \includegraphics[width=0.9\linewidth]{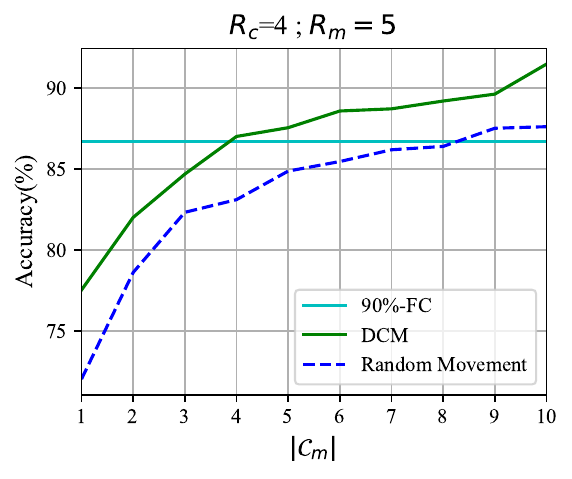}
        \caption{}
        \label{fig:FC_Cm1} 
    \end{subfigure}
    \hfill
    \begin{subfigure}{0.24\textwidth}
        \centering
        \includegraphics[width=0.9\linewidth]{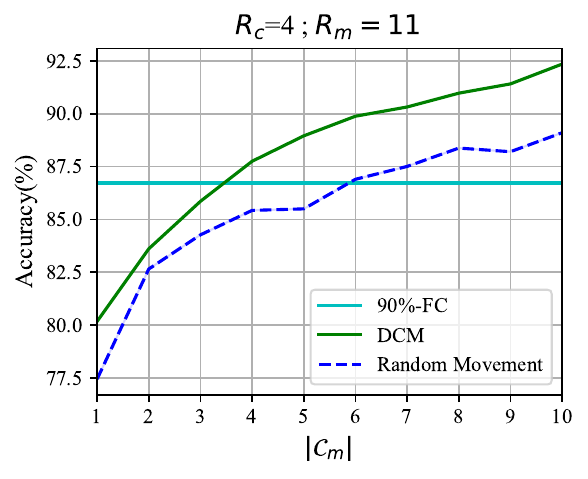}
        \caption{}
        \label{fig:FC_Cm2} 
    \end{subfigure}
    \hfill
    \begin{subfigure}{0.24\textwidth}
        \centering
        \includegraphics[width=0.9\linewidth]{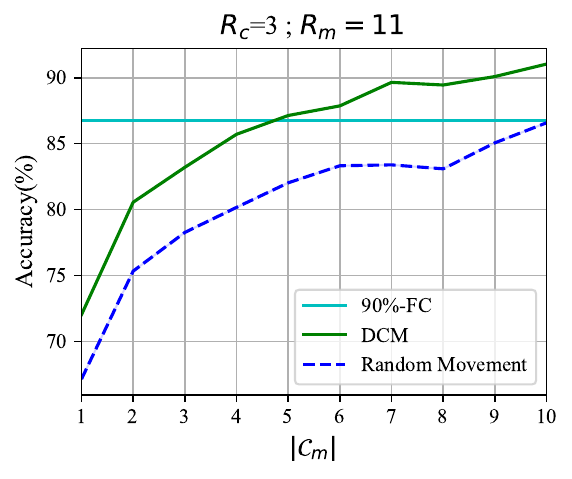}
        \caption{}
        \label{fig:FC_Cm3} 
    \end{subfigure}
    \hfill
    \begin{subfigure}{0.24\textwidth}
        \centering
        \includegraphics[width=0.9\linewidth]{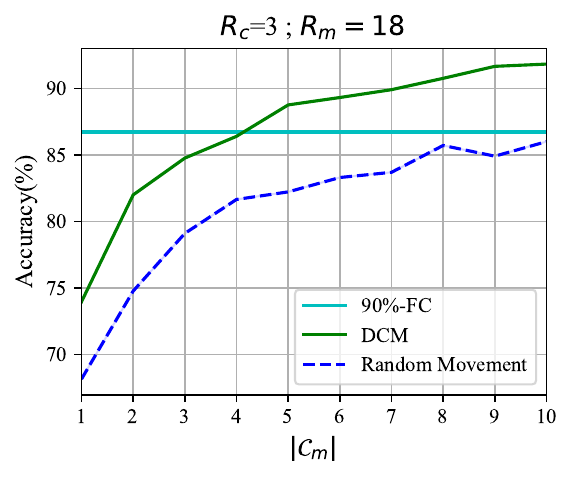}
        \caption{}
        \label{fig:FC_Cm4} 
    \end{subfigure}
    \caption{$|\mathcal{C}_m|$ threshold in a DFL Network.}
    \label{fig:FC_Cm}
\end{figure}

Fig.~\ref{fig:FC_Cm} shows the performance curve with respect to the number of mobile clients. As illustrated in Fig.~\ref{fig:FC_Cm}(\subref{fig:FC_Cm1}) and (\subref{fig:FC_Cm2}), when $R_c=4$, DCM requires to deploy fewer mobile clients in the network to achieve the upper bound compared to Random Movement. Under sparser networks, e.g. $R_c=3$, Fig.~\ref{fig:FC_Cm}(\subref{fig:FC_Cm3}) and (\subref{fig:FC_Cm4}) show that Random Movement fails to achieve the desired performance, even with the highest $R_m$. In contrast, DCM easily reaches the upper bound, utilizing only $|\mathcal{C}_m|=4$ mobile clients. This demonstrates that DCM achieves better deployment efficiency compared to Random Movement, requiring fewer mobile clients in similar configurations to offer adequate performance.

\subsubsection{Joint Impact of Parameters}
So far, we examined the impact of parameters on the DFL performance separately. In this section, the experiments aim to study the joint impact of parameters and the interplay between a pair of parameters and the accuracy of the system. The goal of this section is not to compare different mobility strategies, but rather to examine the impact of network parameters. To this end, we only consider Random Movement as the default and inherent mobility model for DFL systems. Our extensive simulations show that DCM and DAM exhibit similar behavior under these conditions.

\begin{figure}[h]
    \begin{subfigure}{0.24\textwidth}
        \includegraphics[width=0.9\linewidth]{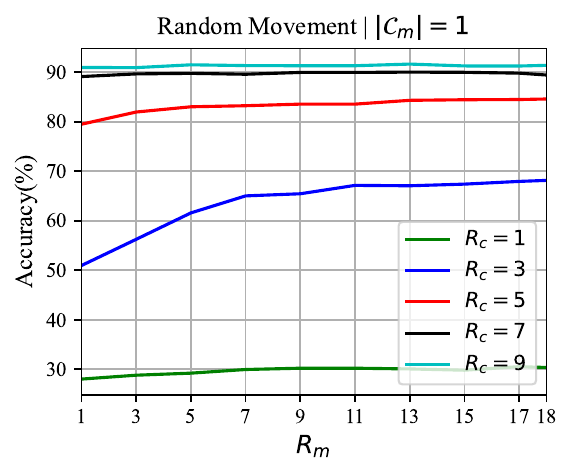}
        \caption{}
        \label{fig:comp1} 
    \end{subfigure}
    \hfill
    \begin{subfigure}{0.24\textwidth}
        \centering
        \includegraphics[width=0.9\linewidth]{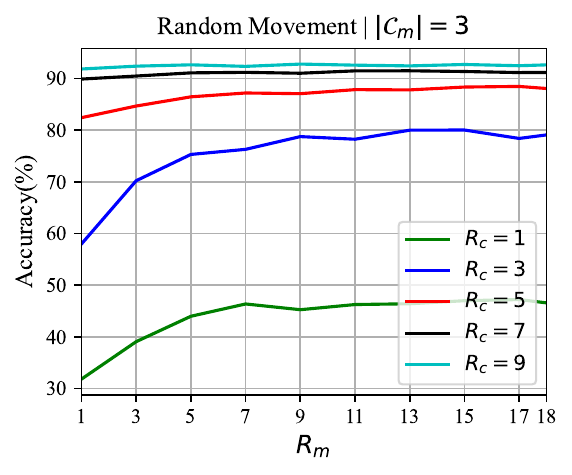} 
        \caption{}
        \label{fig:Comp2} 
    \end{subfigure}
    \hfill
    \begin{subfigure}{0.24\textwidth}
        \centering
        \includegraphics[width=0.9\linewidth]{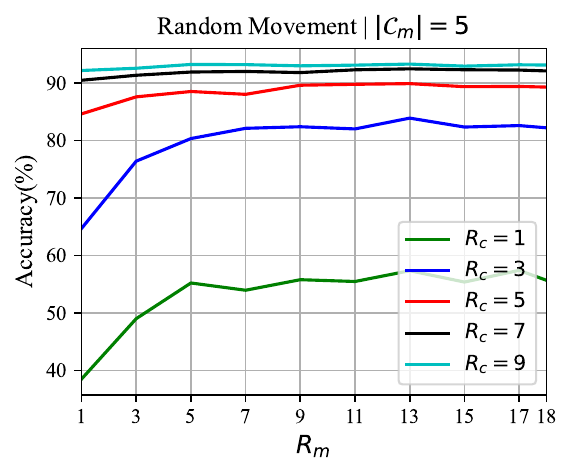}
        \caption{}
        \label{fig:Comp3} 
    \end{subfigure} 
    \hfill
    \begin{subfigure}{0.24\textwidth}
        \centering
        \includegraphics[width=0.9\linewidth]{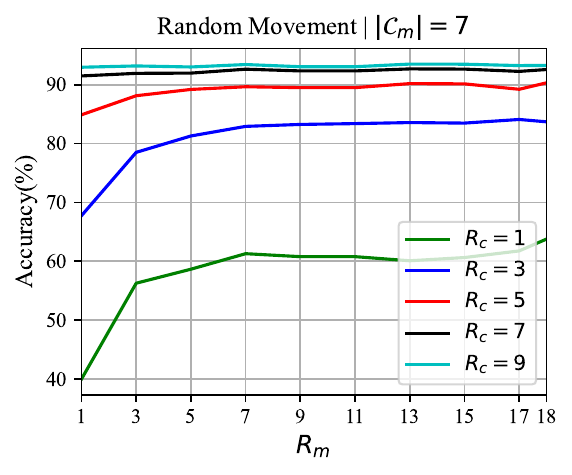}
        \caption{}
        \label{fig:Comp4} 
    \end{subfigure}
    \caption{Joint impact of $R_c$ and $R_m$ with different $|\mathcal{C}_m|$.}
    
    \label{fig:Comp}
\end{figure}

Fig.~\ref{fig:Comp} illustrates how DFL systems with different numbers of mobile clients are affected by variations in communication radius and mobility constraint. The results show that performance saturates beyond a certain threshold of $R_m$, indicating that increasing the speed of mobile clients beyond this point does not directly enhance performance. This threshold value depends on the overall network size. These figures also report that impact of $R_m$ on the DFL system is influenced by $|\mathcal{C}_m|$, as $R_m$ introduces a greater boost in accuracy when more mobile clients are deployed.

Fig.~\ref{fig:Comp} also supports the insights discussed in Section~\ref{sec:theory}, emphasizing the influence of the communication radius. In networks with a large $R_c$, performance remains largely unaffected by the number or speed of mobile clients ($|\mathcal{C}_m|$ and $R_m$), since all regions of the network are already interconnected. In such cases, additional relays are unnecessary for information transfer.

\begin{figure}[h]
    \begin{subfigure}{0.24\textwidth}
        \includegraphics[width=0.9\linewidth]{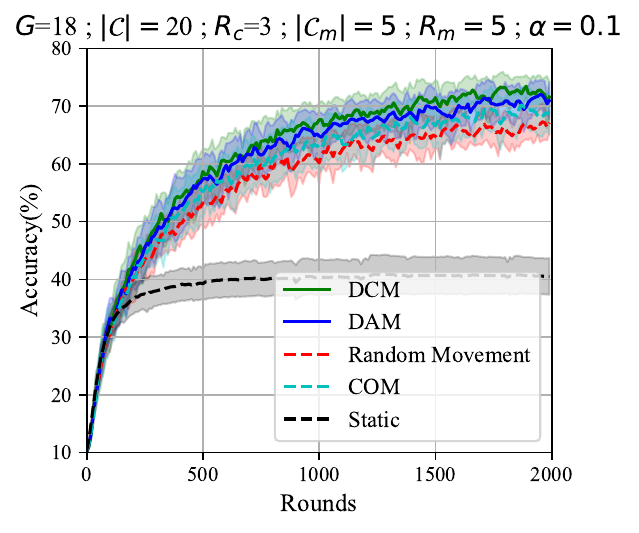}
        \caption{}
        \label{fig:cifar1} 
    \end{subfigure}
    \hfill
    \begin{subfigure}{0.24\textwidth}
        \centering
        \includegraphics[width=0.9\linewidth]{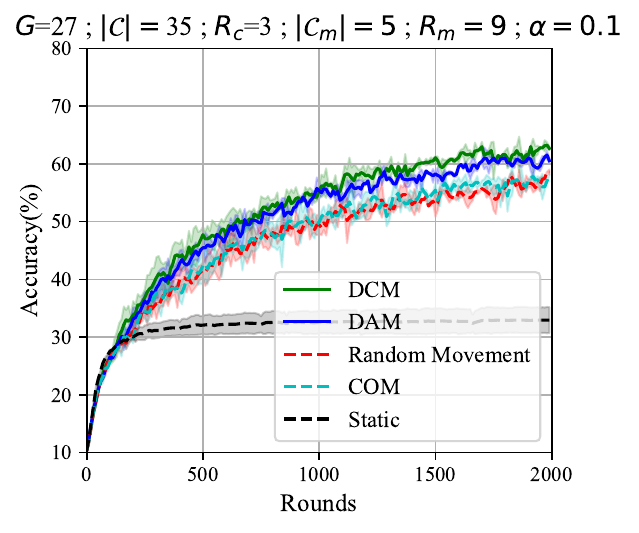}
        \caption{}
        \label{fig:cifar2} 
    \end{subfigure}
    % \label{fig:cif}
    \begin{subfigure}{0.24\textwidth}
        \includegraphics[width=0.9\linewidth]{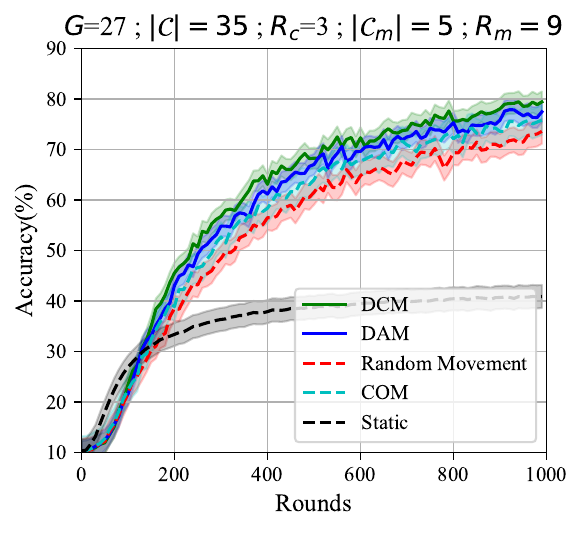}
        \caption{}
        \label{fig:ln1} 
    \end{subfigure}
    \hfill
    \begin{subfigure}{0.24\textwidth}
        \centering
        \includegraphics[width=0.9\linewidth]{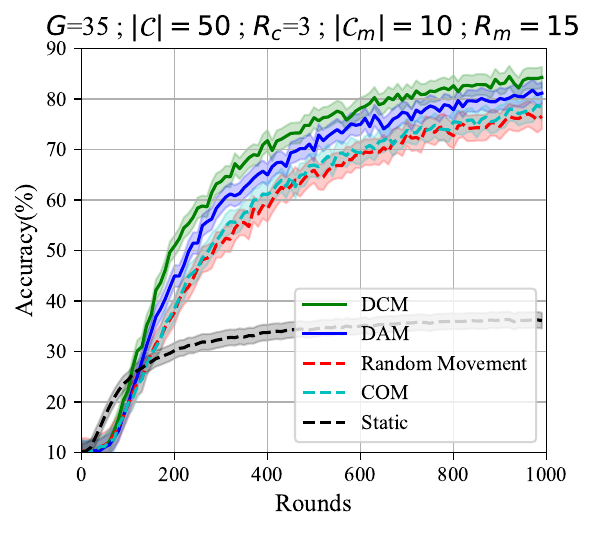} 
        \caption{}
        \label{fig:ln2} 
    \end{subfigure}
    % \caption{DFL performance in larger networks with more clients}
    \caption{Generalization to CIFAR10 dataset and larger networks.}
    \label{fig:ln}
    
\end{figure}

% \begin{figure}[h] 
%     \begin{subfigure}{0.24\textwidth}
%         \includegraphics[width=0.9\linewidth]{Figures/accLn1.pdf}
%         \caption{}
%         \label{fig:ln1} 
%     \end{subfigure}
%     \hfill
%     \begin{subfigure}{0.24\textwidth}
%         \centering
%         \includegraphics[width=0.9\linewidth]{Figures/accLn2.pdf} 
%         \caption{}
%         \label{fig:ln2} 
%     \end{subfigure}
%     \caption{DFL performance in larger networks with more clients}
%     \label{fig:ln}
% \end{figure}

\subsubsection{Generalization}
To validate scalability and robustness of our methods, we extend our experiments to CIFAR-10 dataset and larger networks with more clients. Fig.~\ref{fig:ln}(\subref{fig:cifar1}) and (\subref{fig:cifar2}) illustrate the performance of DFL system in CIFAR-10 classification task in two network configurations, with Fig.~\ref{fig:ln}(\subref{fig:cifar1}) referring to a network of size $18\times18$ with 20 clients and Fig.~\ref{fig:ln}(\subref{fig:cifar2}) referring to a network of size $27\times27$ with 35 clients. These figures show a consistent gain of DCM and DAM over the baselines, illustrating that distribution-aware methods can generalize across different datasets and model architectures.
Moreover, Fig. \ref{fig:ln}(\subref{fig:ln1}) and (\subref{fig:ln2}) demonstrate the superiority of our methodology over the baselines in larger networks based on further extensive simulations on the MNIST dataset with more clients. To further demonstrate the effectiveness of our method, we repeat the experiments in section~\ref{sec:fc} in larger networks.

\begin{figure}[h] 
    \begin{subfigure}{0.24\textwidth}
        \centering
        \includegraphics[width=0.9\linewidth]{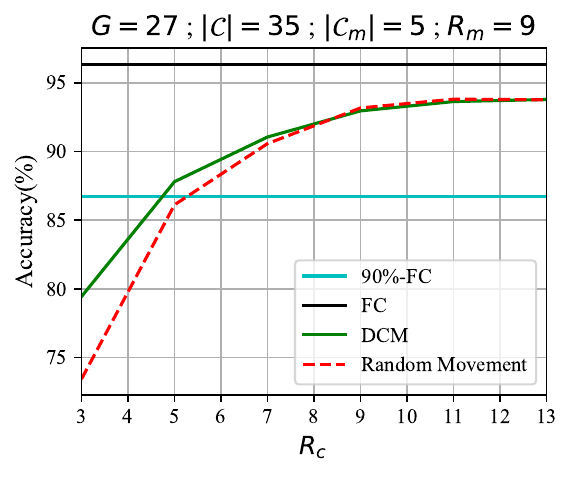} 
        \caption{}
        \label{fig:FC_Rc3} 
    \end{subfigure} 
    \hfill
    \begin{subfigure}{0.24\textwidth}
        \centering
        \includegraphics[width=0.9\linewidth]{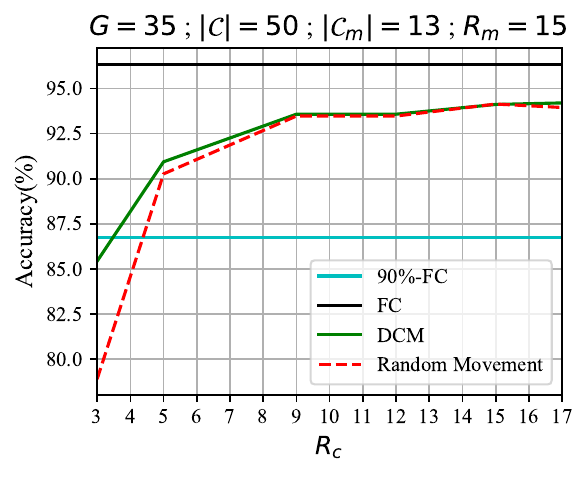} 
        \caption{}
        \label{fig:FC_Rc4} 
    \end{subfigure} 
    \hfill
    \begin{subfigure}{0.24\textwidth}
        \includegraphics[width=0.9\linewidth]{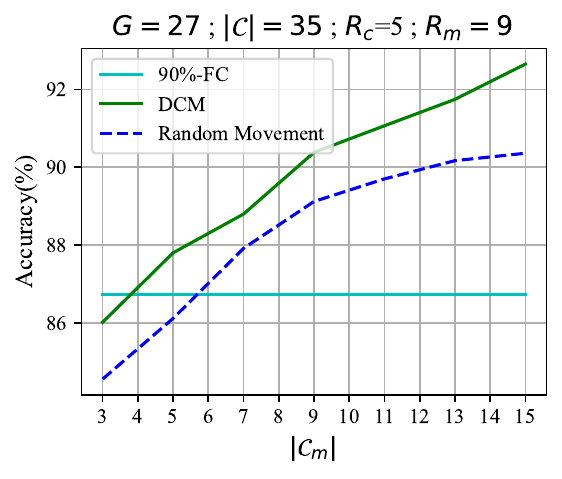}
        \caption{}
        \label{fig:FC_CmL1} 
    \end{subfigure}
    \hfill
    \begin{subfigure}{0.24\textwidth}
        \centering
        \includegraphics[width=0.9\linewidth]{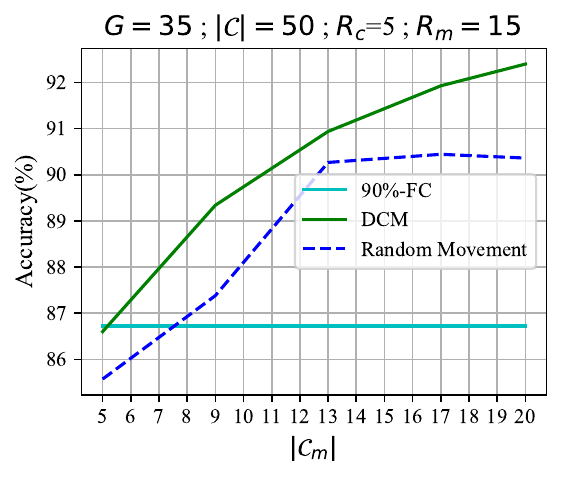}
        \caption{}
        \label{fig:FC_CmL2} 
    \end{subfigure}
    \caption{Threshold values for $R_c$ and $|\mathcal{C}_m|$ in a larger networks.}
    \label{fig:FC_LL}
\end{figure}

Fig.~\ref{fig:FC_LL} illustrates the performance of DFL system in two larger networks with $35$ and $50$ clients, respectively, in networks of size $27\times27$ and $35\times35$, demonstrating the generalization and robustness of our method to larger networks while maintaining the consistent gain over the baseline in different configurations. Moreover, Fig.~\ref{fig:FC_LL}(\subref{fig:FC_Rc3}) and (\subref{fig:FC_Rc4}) extend the communication efficiency feature of DCM to larger networks and more complex systems. Similarly, Fig.~\ref{fig:FC_LL}(\subref{fig:FC_CmL1}) and (\subref{fig:FC_CmL2}) demonstrate that DCM requires fewer mobile clients to provide the desired accuracy in these networks, confirming the superiority of DCM deployment efficiency over the baseline.

\subsubsection{Empirical Complexity Reduction of DCM}
To quantify the computational benefit of DCM, we report $\mathbb{E}[|\mathcal{L}_c|]$ over 10 Monte Carlo trials for multiple network settings. Since DAM and the COM baseline evaluate all $G^2$ grid locations while DCM evaluates only $|\mathcal{L}_c|$ centers, the empirical reduction factor is $\frac{G^2}{\mathbb{E}[|\mathcal{L}_c|]}$. As shown in Table~\ref{tab:complexity_scaling}, DCM reduces the candidate search space across the tested configurations, with larger $R_c$ typically decreasing $\mathbb{E}[|\mathcal{L}_c|]$ and thus increasing the reduction factor. 

\begin{table}[h]
\centering
\caption{Empirical candidate search complexity scaling.}
\label{tab:complexity_scaling}
\renewcommand{\arraystretch}{1.2}
\setlength{\tabcolsep}{4pt}
\begin{tabular}{|c|c|c|c|}
\hline
Method & DCM & COM \& DAM & Reduction Factor \\
\hline
\textbf{Setting} $(G, |\mathcal{C}|, |\mathcal{C}_m|, R_c)$ 
& $\mathbb{E}[|\mathcal{L}_c|]$ 
& $G^2$ 
& $\frac{G^2}{\mathbb{E}[|\mathcal{L}_c|]}$ \\
\hline
$(18, 20, 3, 3)$  & 7.0  & 324  & 46.29 \\
\hline
$(27, 35, 5, 3)$  & 13.3 & 729  & 54.81 \\
% \hline
% $(27, 35, 5, 5)$  & 7.9  & 729  & 92.28 \\
% \hline
% $(35, 50, 10, 3)$ & 19.2 & 1225 & 63.80 \\
\hline
$(35, 50, 10, 5)$ & 12.0 & 1225 & 102.08 \\
\hline
\end{tabular}
\end{table}

\subsubsection{Empirical Evaluation of Connectivity Parameter $B$}

To further examine the interaction between mobility strategies and the connectivity parameter $B$ in Theorem~\ref{th:theorem}, we empirically evaluate $B$ under different mobility patterns with $|\mathcal{C}|=20$ total clients and $|\mathcal{C}_m|=1$ mobile client under $R_c=3$ and $R_m=5$. In our experiments, $B$ is approximately measured as the number of communication rounds required until all static clients establish at least one connection with the mobile client. Table~\ref{tab:B_eval} reports the $\mathbb{E}[B]$ over 10 Monte Carlo trials.

\begin{table}[h]
\centering
\caption{Empirical evaluation of connectivity parameter $B$.}
\label{tab:B_eval}
\renewcommand{\arraystretch}{1.2}
\setlength{\tabcolsep}{6pt}
\begin{tabular}{|c|c|}
\hline
\textbf{Mobility Strategy} & $\mathbb{E}[B]$ \\
\hline
DCM & 34.6 \\
\hline
DAM & 68.2 \\
\hline
Random Movement & 139.9 \\
\hline
COM & 75.1 \\
\hline
\end{tabular}
\end{table}

The results confirm our theoretical intuitions. DCM achieves the smallest $B$ due to its structured movement pattern. DAM and COM exhibit similar behavior and Random Movement yields the largest $B$ due to unstructured exploration.

% , indicating faster temporal connectivity and more efficient information mixing across the network. While DAM reduces heterogeneity through distribution-aware exploration, its unrestricted search results in a larger $B$ compared to DCM. COM improves instantaneous connectivity but does not consistently minimize global mixing time, and Random Movement yields the largest $B$ due to unstructured exploration. These observations demonstrate that DCM implicitly aligns improvements in both the heterogeneity term $\hat{\tau}$ and the connectivity parameter $B$, reinforcing the convergence behavior predicted by Theorem~\ref{th:theorem}.

\subsubsection{Empirical Validation of Design Choice}

To empirically validate the effectiveness of our design choices and support our justifications in Section~\ref{sec:framework}, we report the final accuracy of DFL under $R_c=3, |\mathcal{C}_m|=3,R_m=5$ in a network with 20 clients in table \ref{tab:soft_norm} over 6 Monte Carlo trials. 
This empirical evaluation demonstrates that utilizing normalization in our methodology results in a better performance compared to using Softmax function for generating movement probabilities.

\begin{table}[h]
\centering
\caption{Normalization/Softmax for DCM/DAM. \vspace{-2mm}}
\begin{tabular}{|c|c|c|}
\hline
\textbf{Movement Pattern} & Softmax & Normalization \\
\hline
DCM & $77.80\%\pm2.28$\% & $80.83\%\pm1.34$\% \\
\hline
DAM & $77.24\%\pm3.46$\% & $79.85\% \pm2.68$\% \\
\hline
\end{tabular}
\label{tab:soft_norm} \vspace{-2mm}
\end{table}

\vspace{-4mm}

\section{Conclusions and Future Works}

\noindent In this work, we considered the problem of performance degradation of DFL systems in sparse networks with limited connectivity among the clients under heterogeneous data distribution. We derived the convergence bound in the DFL system under mobility, illustrating the impact of user mobility on learning performance. Following the intuition of information transfer based on the theoretical results in this work, we proposed a DFL framework that utilizes two novel mobility strategies to resolve data heterogeneity. Through extensive simulations, we validated the superiority of our proposed framework over the baseline, in addition to conducting a comprehensive analysis of the impact of network parameters on DFL performance.

\section*{Acknowledgment}

The authors would like to thank Dr. Aritra Mitra for his helpful discussions on DFL convergence under mobility.

\newpage

\bibliographystyle{IEEEtran}
\bibliography{references}

\appendix

\subsection{Proof of Lemma \ref{lem:stack}}
\vspace{-5mm}
\begin{align}
\mathbf{X}^{(t)} 
&= \big(\mathbf{X}^{(t-1)} - \eta \mathbf{G}^{(t-1)}\big) \mathbf{W}^{(t-1)} \notag \\
&= \mathbf{X}^{(t-1)} \mathbf{W}^{(t-1)} - \eta \mathbf{G}^{(t-1)} \mathbf{W}^{(t-1)} \notag \\
&= \big(\mathbf{X}^{(t-2)} - \eta \mathbf{G}^{(t-2)}\big) \mathbf{W}^{(t-2)} \mathbf{W}^{(t-1)} 
    - \eta \mathbf{G}^{(t-1)} \mathbf{W}^{(t-1)} \notag \\
&= \mathbf{X}^{(t-2)} \mathbf{W}^{(t-2)} \mathbf{W}^{(t-1)}
    - \eta \mathbf{G}^{(t-2)} \mathbf{W}^{(t-2)} \mathbf{W}^{(t-1)}  \notag \\& 
\quad - \eta \mathbf{G}^{(t-1)} \mathbf{W}^{(t-1)} \notag \\
&= \big(\mathbf{X}^{(t-3)} - \eta \mathbf{G}^{(t-3)}\big) \mathbf{W}^{(t-3)} \mathbf{W}^{(t-2)} \mathbf{W}^{(t-1)} \notag \\
&\quad - \eta \mathbf{G}^{(t-2)} \mathbf{W}^{(t-2)} \mathbf{W}^{(t-1)}
    - \eta \mathbf{G}^{(t-1)} \mathbf{W}^{(t-1)} \notag \\
&= \mathbf{X}^{(t-3)} \mathbf{W}^{(t-3)} \mathbf{W}^{(t-2)} \mathbf{W}^{(t-1)} \notag \\
&\quad - \eta \mathbf{G}^{(t-3)} \mathbf{W}^{(t-3)} \mathbf{W}^{(t-2)} \mathbf{W}^{(t-1)} \notag \\
&\quad - \eta \mathbf{G}^{(t-2)} \mathbf{W}^{(t-2)} \mathbf{W}^{(t-1)}
    - \eta \mathbf{G}^{(t-1)} \mathbf{W}^{(t-1)} . \notag
\end{align}

Resuming the same procedure, we introduce the closed form for representing $\mathbf{X}^{(t)}$ based on the previous $B$ steps:
\begin{equation}\label{x(t)}
\mathbf{X}^{(t)} = \mathbf{X}^{(t-B)} \prod_{k=t-B}^{t-1} \mathbf{W}^{(k)}  \notag
- \eta \sum_{k=t-B}^{t-1} \left( \mathbf{G}^{(k)} \prod_{j=k}^{t-1} \mathbf{W}^{(j)} \right).
\end{equation}

\subsection{Proof of Corollary \ref{th:corol2}}

To prove Corollary \ref{th:corol2}, we need to refer to Corollary 2 in ~\cite{nedic2014distributed}. In this work, the authors prove $|[\mathbf{A}(t:t+B)]_{ij}-\phi|\leq C\lambda^{B}$, where $\mathbf{A}(t:t+B)=\mathbf{A}(t)\mathbf{A}(t+1)...\mathbf{A}(t+B)$ and $\phi\geq\frac{\delta}{N}$ is a scalar, where $\delta \geq \frac{1}{N^{NB}}$, and $\lambda=(1-\frac{1}{N^{NB}})^{\frac{1}{B}}$. 

Plugging $\boldsymbol{\psi}^{(B)}=\prod_{t=kB}^{(k+1)B-1}\mathbf{W}^{(t)}$ and $\phi=\frac{1}{N}$, we can derive the following:

\begin{align}
    |\boldsymbol{\psi}^{B}_{ij}-\frac{1}{N}|\leq C\lambda^{B} \notag
\end{align}

Therefore, we can show:
\begin{align}
    \| \boldsymbol{\psi}^{(B)} - \frac{\mathbf{11}^\top}{N} \|_F^2 \leq C^2N^2\lambda^{2B} \notag
\end{align}

\subsection{Proof of Theorem \ref{th:theorem}}
Thanks to Lemma \hyperref[lem:lemm2]{2} and Lemma \hyperref[lem:lemm3]{3}, we can derive the final convergence bound on the optimization error for the DFL system operating under mobility with the assumption of convexity of all local objective functions, where $\varepsilon^{(m)}=[\xi_1^{(t)},\xi_2^{(t)},...,\xi_N^{(t)}]$.

\begin{align}
    &f(\bar{\mathbf{x}}^{(t)}) - f^*  \leq \frac{1}{2\eta(1-2L\eta)} \notag
    \\& \Bigg( 
\|\bar{\mathbf{x}}^{(t)} - \mathbf{x}^*\|^2 \notag
- \mathbb{E}_{\varepsilon^{(t)}|\{\varepsilon^{(m)}\}_{m=0}^{t-1}}
\Big\{\|\bar{\mathbf{x}}^{(t+1)} - \mathbf{x}^*\|^2\Big\}  \notag
\\&+ \frac{\eta^2 \bar{\sigma}^2}{N}
+ \frac{L\eta}{N}(2L\eta+1)\|\mathbf{X}^{(t)} - \bar{\mathbf{X}}^{(t)}\|_F^2
\Bigg) \notag
\end{align}

\begin{align}
    & \mathbb{E}\big\{ f(\bar{\mathbf{x}}^{(t)}) - f^* \big\} \notag
\leq \\&\frac{1}{2\eta(1-2L\eta)} \Bigg(
\mathbb{E}\|\bar{\mathbf{x}}^{(t)} - \mathbf{x}^*\|^2     \notag
- \mathbb{E}\|\bar{\mathbf{x}}^{(t+1)} - \mathbf{x}^*\|^2
\\&+ \frac{\eta^2 \bar{\sigma}^2}{N}
+ \frac{L\eta}{N}(2L\eta+1)\, \mathbb{E}\|\mathbf{X}^{(t)} - \bar{\mathbf{X}}^{(t)}\|_F^2
\Bigg)   \notag
\end{align}

Taking the summation over all rounds, Theorem \ref{th:theorem} is proved.

\begin{align}
&\frac{1}{T+1}\sum_{t=0}^T \mathbb{E}\big\{ f(\bar{\mathbf{x}}^{(t)}) - f^* \big\}
\leq \frac{1}{2\eta(1-2L\eta)   \notag
}\\ & \Bigg(  \nonumber
\frac{\|\bar{\mathbf{x}}^{(0)} - \mathbf{x}^*\|^2}{T+1}
+ \frac{\eta^2 \bar{\sigma}^2}{N}   \nonumber
+ \frac{L\eta^3}{N}(2L\eta+1)A_0         \nonumber
\Bigg),
\end{align}

where we define $A_0 = \left(1 + \frac{2}{p}\right)\left(\frac{2}{p}\right)\hat{\tau}\big(\frac{(1-p)}{p} + C^2 N^2 \lambda^{2B}\big)$.

% \begin{align}
%     A_0 = \left(1 + \frac{2}{p}\right)\left(\frac{2}{p}\right)\hat{\tau}\big(\frac{(1-p)}{p} + C^2 N^2 \lambda^{2B}\big). \notag
% \end{align}

\subsection{Needed Lemmas}
In this section, we derive the proof for the needed lemmas, with Lemma \ref{lem:lemm2} showing the descent flow in DFL and Lemma \ref{lem:lemm3} showing the consensus control. We define $\mathcal{Z}=\{\varepsilon^{(m)}\}_{m=0}^{t-1}$,

\begin{lemma}
[Descent Lemma]\label{lem:lemm2}
\begin{align} \nonumber
\mathbb{E} \| \bar{\mathbf{x}}^{(t+1)} - \mathbf{x}^* \|^2
\leq 
&\| \bar{\mathbf{x}}^{(t)} - \mathbf{x}^*\|^2
+ \frac{\eta_t^2}{N} \bar{\sigma}^2
\\& + 2\eta_t (2L\eta_t - 1)\big(f(\bar{\mathbf{x}}^{(t)}) - f^*\big) \nonumber
\\&+ \frac{L\eta_t}{N} (2L\eta_t + 1) \| \bar{\mathbf{X}}^{(t)} - \mathbf{X}^{(t)} \|_F^2 \nonumber
\end{align}
\end{lemma}
% In this section, we establish the following lemma showing the descent flow in DFL. We define $\mathcal{Z}=\{\varepsilon^{(m)}\}_{m=0}^{t-1}$, 

Proof.

Our proof for this lemma closely follows the steps in \cite{koloskova2020unified}.
\begin{align}
    &\left\| \bar{\mathbf{x}}^{(t+1)} - \mathbf{x}^* \right\|^2
= \left\| \bar{\mathbf{x}}^{(t)} - \frac{\eta_t}{N} \sum_i \mathbf{g}_i^{(t)} - \mathbf{x}^* \right\|^2 \notag
\end{align}

\begin{align}
   =  &\| \bar{\mathbf{x}}^{(t)} - \mathbf{x}^* - \frac{\eta_t}{N} \sum_i \nabla f_i \left(\mathbf{x}_i^{(t)}\right) \nonumber  \\ 
&+ \frac{\eta_t}{N} \sum_i \nabla f_i \left(\mathbf{x}_i^{(t)}\right)
- \frac{\eta_t}{N} \sum_i \mathbf{g}_i^{(t)} \|^2 \nonumber
\end{align}

\begin{align}
    =&\left\| \bar{\mathbf{x}}^{(t)} - \mathbf{x}^* - \frac{\eta_t}{N} \sum_i \nabla f_i \left(\mathbf{x}_i^{(t)}\right) \right\|^2  \notag
\\&+ \frac{\eta_t^2}{N^2} \left\| \sum_i \nabla f_i(\mathbf{x}_i^{(t)}) - \sum_i \mathbf{g}_i^{(t)} \right\|^2   \notag
\\&+ 2 \frac{\eta_t}{N} \Big\langle \bar{\mathbf{x}}^{(t)} - \mathbf{x}^* - \frac{\eta_t}{N} \sum_i \nabla f_i(\mathbf{x}_i^{(t)}), \notag \\& \notag
\sum_i \nabla f_i(\mathbf{x}_i^{(t)}) - \sum_i \mathbf{g}_i^{(t)} \Big\rangle \notag
\end{align}

\begin{align}
    T_3 = \Big\langle \bar{\mathbf{x}}^{(t)} - \mathbf{x}^* - \frac{\eta_t}{N} \sum_i \nabla f_i(\mathbf{x}_i^{(t)}), 
\sum_i \nabla f_i(\mathbf{x}_i^{(t)}) - \sum_i \mathbf{g}_i^{(t)} \Big\rangle \notag
\end{align}

Using $\mathbb{E} \ \mathbf{g}_i^{(t)}=\mathbb{E} \ \nabla F_i(\mathbf{x}_i^{(t)},\xi_i^{(t)})=\nabla f_i(\mathbf{x}_i^{(t)})$,

\begin{equation}
    \mathbb{E}_{\varepsilon^{(t)} |\mathcal{Z}} T_3 =0 \notag
\end{equation}
    
\begin{align}
    & T_2 = \frac{\eta_t^2}{N^2} \left\| \sum_i \nabla f_i(\mathbf{x}_i^{(t)}) - \sum_i \mathbf{g}_i^{(t)} \right\|^2 \notag
\end{align}

Taking the expectation and using Assumption \ref{assump:var},

\begin{align}
    \mathbb{E}_{\varepsilon^{(t)}|\mathcal{Z}} T_2  \notag
\leq &
\frac{\eta_t^2}{N^2} \sum_i \mathbb{E} \left\| \nabla f_i(\mathbf{x}_i^{(t)}) - \nabla F_i(\mathbf{x}_i^{(t)}, \xi_i^{(t)}) \right\|^2   \notag
\leq \\&\frac{\eta_t^2}{N^2} \sum_i \sigma_i^2
= \frac{\eta_t^2}{N} \bar{\sigma}^2   \notag
\end{align}

\begin{align}
    T_1 =& \left\| \bar{\mathbf{x}}^{(t)} - \mathbf{x}^* - \frac{\eta_t}{N} \sum_i \nabla f_i(\mathbf{x}_i^{(t)}) \right\|^2 \notag
=\\& \|\bar{\mathbf{x}}^{(t)} - \mathbf{x}^*\|^2 + \eta_t^2 \left\| \frac{1}{N} \sum_i \nabla f_i(\mathbf{x}_i^{(t)}) \right\|^2 \notag
\\& -2\eta_t \Big\langle \bar{\mathbf{x}}^{(t)} - \mathbf{x}^*, \frac{1}{N} \sum_i \nabla f_i(\mathbf{x}_i^{(t)}) \Big\rangle   \notag
\end{align}

\begin{align}
&B_1 = \left\| \frac{1}{N} \sum_i \nabla f_i(\mathbf{x}_i^{(t)}) \right\|^2 \notag
= \\&\left\| \frac{1}{N} \sum_i \nabla f_i(\mathbf{x}_i^{(t)}) - \nabla f_i(\bar{\mathbf{x}}^{(t)}) \notag
+ \nabla f_i(\bar{\mathbf{x}}^{(t)}) - \nabla f_i(\mathbf{x}^*) \right\|^2
\end{align}

Using Property \hyperref[prop:properties]{2},

\begin{align}
    B_1 \leq 2  \left\|\frac{1}{N} \sum_i \nabla f_i(\mathbf{x}_i^{(t)}) - \nabla f_i(\bar{\mathbf{x}}^{(t)}) \right\|^2 \notag
\\+ 2  \left\|\frac{1}{N} \sum_i \nabla f_i(\bar{\mathbf{x}}^{(t)}) - \nabla f_i(\mathbf{x}^*) \right\|^2 \notag
\end{align}

\begin{align}
    = \frac{2}{N^2}  \|\sum_{i=1}^{N}\nabla f_i(\mathbf{x}_i^{(t)}) - \nabla f_i(\bar{\mathbf{x}}^{(t)})\|^2 \notag
\\+ \frac{2}{N^2}  \|\sum_{i=1}^{N}\nabla f_i(\bar{\mathbf{x}}^{(t)}) - \nabla f_i(\mathbf{x}^*)\|^2 \notag
\end{align}

\begin{align}
    \leq \frac{2}{N} \sum_{i=1}^{N} \|\nabla f_i(\mathbf{x}_i^{(t)}) - \nabla f_i(\bar{\mathbf{x}}^{(t)})\|^2   \notag
 \\
 + \frac{2}{N} \sum_{i=1}^{N} \|\nabla f_i(\bar{\mathbf{x}}^{(t)}) - \nabla f_i(\mathbf{x}^*)\|^2    \notag
\end{align}

Using Property \hyperref[prop:properties]{1} powered by Assumption \ref{assump:convx},

\begin{align}
    B_1 \leq \frac{2}{N} \sum_i L^2 \|\mathbf{x}_i^{(t)} - \bar{\mathbf{x}}^{(t)}\|^2 \notag
\end{align}

\begin{align}
 + \frac{2}{N} \sum_i \Big[ 2L\big(f_i(\bar{\mathbf{x}}^{(t)}) - f_i(\mathbf{x}^*)\big) 
- \langle \nabla f_i(\mathbf{x}^*), \bar{\mathbf{x}}^{(t)} - \mathbf{x}^* \rangle \Big]  \notag
\end{align}

\begin{align}
    = \frac{2L^2}{N} \sum_i \|\mathbf{x}_i^{(t)} - \bar{\mathbf{x}}^{(t)}\|^2
+ 4L \big(f(\bar{\mathbf{x}}^{(t)}) - f(\mathbf{x}^*)\big)  \notag
\end{align}

\begin{align}
    B_2 =& -\frac{2\eta_t}{N} \left\langle \bar{\mathbf{x}}^{(t)} - \mathbf{x}^*, \sum_i \nabla f_i(\mathbf{x}_i^{(t)}) \right\rangle   \notag
\\= &-\frac{2\eta_t}{N} \sum_i \langle \bar{\mathbf{x}}^{(t)} - \mathbf{x}^*, \nabla f_i(\mathbf{x}_i^{(t)}) \rangle \notag
\end{align}

\begin{align}
    =& -\frac{2\eta_t}{N} \sum_i \langle \bar{\mathbf{x}}^{(t)} - \mathbf{x}_i^{(t)}, \nabla f_i(\mathbf{x}_i^{(t)}) \rangle \notag
\\& + \langle \mathbf{x}_i^{(t)} - \mathbf{x}^*, \nabla f_i(\mathbf{x}_i^{(t)}) \rangle \notag
\end{align}

Using Property \hyperref[prop:properties]{1} powered by Assumptions \ref{assump:smooth} and \ref{assump:convx},

\begin{align}
    B_2 \leq -\frac{2\eta_t}{N} \sum_i \Big[ f_i(\bar{\mathbf{x}}^{(t)}) - f_i(\mathbf{x}_i^{(t)})    \notag
    \\- \frac{L}{2}\|\bar{\mathbf{x}}^{(t)} - \mathbf{x}_i^{(t)}\|^2
+ f_i(\mathbf{x}_i^{(t)}) - f_i(\mathbf{x}^*) \Big]       \notag
\end{align}

\begin{align}
    = -2\eta_t \big(f(\bar{\mathbf{x}}^{(t)}) - f(\mathbf{x}^*)\big)
+ \frac{L\eta_t}{N} \sum_i \|\bar{\mathbf{x}}^{(t)} - \mathbf{x}_i^{(t)}\|^2   \notag
\end{align}

\begin{align}
    = -2\eta_t \big(f(\bar{\mathbf{x}}^{(t)}) - f(\mathbf{x}^*)\big)
+ \frac{L\eta_t}{N} \| \bar{\mathbf{X}}^{(t)} - \mathbf{X}^{(t)} \|_F^2       \notag
\end{align}

Therefore, the descent lemma can be derived as follows:

\begin{align}
     \mathbb{E} \| \bar{\mathbf{x}}^{(t+1)} - \mathbf{x}^* \|^2
\leq\|& \bar{\mathbf{x}}^{(t)} - \mathbf{x}^*\|^2   \notag
+ \frac{\eta_t^2}{N} \bar{\sigma}^2
\\&+ 2\eta_t (2L\eta_t - 1)\big(f(\bar{\mathbf{x}}^{(t)}) - f^*\big) \notag
\\& + \frac{L\eta_t}{N} (2L\eta_t + 1) \| \bar{\mathbf{X}}^{(t)} - \mathbf{X}^{(t)} \|_F^2   \notag
\end{align}

\begin{lemma}
[Consensus Control]\label{lem:lemm3}
\begin{align}
    &\|\mathbf{X}^{(t)} - \bar{\mathbf{X}}^{(t)}\|_F^2   \notag
\leq \\& \left(1 + \frac{2}{p}\right)\frac{2}{p}\,\eta^2 \hat{\tau}\Big(\frac{(1-p)}{p} + C^2 N^2 \lambda^{2B}\Big)   \notag
\end{align}
\end{lemma}

Proof.

\begin{align}
    \|\bar{\mathbf{X}}^{(t)} - \mathbf{X}^{(t)}\|_F^2
= \left\| \mathbf{X}^{(t)} \left(\mathbf{I} - \frac{\mathbf{11}^\top}{N}\right) \right\|_F^2 \notag
\end{align}

Thanks to Lemma \ref{lem:stack}, we can define 
\begin{align}
    \boldsymbol{\psi}^{(B)}= \prod_{k=t-B}^{t-1} \mathbf{W}^{(k)} \notag
    \\ \mathbf{D}_B=\sum_{k=t-B}^{t-1} \big[ \mathbf{G}^{(k)} \prod_{j=k}^{t-1} \mathbf{W}^{(j)}\big] \notag
\end{align}

We will further insert them and resume the derivation as follows.

\begin{align}
    \|\bar{\mathbf{X}}^{(t)} - \mathbf{X}^{(t)}\|_F^2= \left\| \Big((\mathbf{X}^{(t-B)} \boldsymbol{\psi}^{(B)}) - \eta \mathbf{D}_B\Big)
\left(\mathbf{I} - \frac{\mathbf{11}^\top}{N}\right) \right\|_F^2   \notag
\\= \left\| \mathbf{X}^{(t-B)}\boldsymbol{\psi}^{(B)} - \bar{\mathbf{X}}^{(t-B)} - \eta \mathbf{D}_B\left(\mathbf{I} - \frac{\mathbf{11}^\top}{N}\right) \right\|_F^2 \notag
\end{align}

Using Property \hyperref[prop:properties]{2}, 
\begin{align}
     \|\bar{\mathbf{X}}^{(t)} - \mathbf{X}^{(t)}\|_F^2\leq (1+\alpha) \|\mathbf{X}^{(t-B)}\boldsymbol{\psi}^{(B)} - \bar{\mathbf{X}}^{(t-B)}\|_F^2   \notag
\\+ (1+\alpha^{-1}) \eta^2 \|\mathbf{D}_B \left(\mathbf{I} - \frac{\mathbf{11}^\top}{N}\right)\|_F^2   \notag
\end{align}

Inserting $\alpha=\frac{p}{2}$, and using Assumption \ref{assump:mixing},

\begin{align}
   \|\bar{\mathbf{X}}^{(t)} - \mathbf{X}^{(t)}\|_F^2 \leq & \left(1+\frac{p}{2}\right)(1-p)\|\mathbf{X}^{(t-B)} - \bar{\mathbf{X}}^{(t-B)}\|_F^2   \notag
\\&+ \left(1+\frac{2}{p}\right)\eta^2 \|\mathbf{D}_B \left(\mathbf{I} - \frac{\mathbf{11}^\top}{N}\right)\|_F^2   \notag
\\\leq& \left(1-\frac{p}{2}\right)\|\mathbf{X}^{(t-B)} - \bar{\mathbf{X}}^{(t-B)}\|_F^2 \notag
\\&+ \left(1+\frac{2}{p}\right)\eta^2 \|\mathbf{D}_B \left(\mathbf{I} - \frac{\mathbf{11}^\top}{N}\right)\|_F^2   \notag
\end{align}

Before we proceed to bounding the remaining terms, we introduce a simple short lemma.

\begin{align}
    &\left\| \mathbf{G}^{(t)} \left(\mathbf{W}^{(t)} - \frac{\mathbf{11}^\top}{N}\right) \right\|_F^2= \notag
\\& \left\| \mathbf{G}^{(t)} - \bar{\mathbf{g}}^{(t)} 1^\top \right\|_F^2
\left\| \mathbf{W}^{(t)} - \frac{\mathbf{11}^\top}{N} \right\|_F^2
\leq \hat{\tau}(1-p)    \notag
\end{align}

Where $\mathbf{W}^{(t)}$ can be any doubly stochastic mixing matrix  and $\bar{\mathbf{g}}^{(t)}$ and $\|\mathbf{G}^{(t)}-\bar{\mathbf{g}}^{(t)}1^\top\|_F^2$ are bounded as follows:

\begin{align}
    &\bar{\mathbf{g}}^{(t)} = \sum_i \mathbf{G}^{(t)}_{[:,i]} \notag \\&
    \|\mathbf{G}^{(t)} - \bar{\mathbf{g}}^{(t)} 1^\top\|_F^2
= \sum_{i=1}^N \|\mathbf{g}_i^{(t)} - \bar{\mathbf{g}}^{(t)}\|^2
\leq \sum_{i=1}^N \tau_i = \hat{\tau}   \notag
\end{align}

We resume the derivation as follows.

\begin{align}
    &T_4 = \left\| \mathbf{D}_B \left(\mathbf{I} - \frac{\mathbf{11}^\top}{N}\right) \right\|_F^2   \notag
= \\&
\left\| \sum_{k=t-B}^{t-1} \Big[ \mathbf{G}^{(k)} \prod_{j=k}^{t-1} \mathbf{W}^{(j)} \Big]
\left(\mathbf{I} - \frac{\mathbf{11}^\top}{N}\right) \right\|_F^2  \notag
\end{align}

\begin{align}
    \leq \sum_{k=t-B}^{t-1} \left\| \mathbf{G}^{(k)} \prod_{j=k}^{t-1} \mathbf{W}^{(j)}
\left(\mathbf{I} - \frac{\mathbf{11}^\top}{N}\right) \right\|_F^2   \notag
\end{align}

\begin{align}
    = &\left\| \mathbf{G}^{(t-1)} \mathbf{W}^{(t-1)} \left(\mathbf{I} - \frac{\mathbf{11}^\top}{N}\right) \right\|_F^2  \notag
 \\& + \left\| \mathbf{G}^{(t-2)} \mathbf{W}^{(t-2)} \mathbf{W}^{(t-1)} \left(\mathbf{I} - \frac{\mathbf{11}^\top}{N}\right) \right\|_F^2    \notag
\\& + \cdots    \notag
\\&+ \left\| \mathbf{G}^{(t-B)} \boldsymbol{\psi}^{(B)} \left(\mathbf{I} - \frac{\mathbf{11}^\top}{N}\right) \right\|_F^2   \notag
\\=& \sum_{k=t-B+1}^{t-1} \left\| \mathbf{G}^{(k)} \prod_{j=k}^{t-1} \mathbf{W}^{(j)}
\left(\mathbf{I} - \frac{\mathbf{11}^\top}{N}\right) \right\|_F^2   \notag \\& + \left\| \mathbf{G}^{(t-B)} \boldsymbol{\psi}^{(B)} \left(\mathbf{I} - \frac{\mathbf{11}^\top}{N}\right) \right\|_F^2   \notag
\end{align}

Separating the product terms using Property \hyperref[prop:properties]{3} gives us:

\begin{align}
    \sum_{k=t-B+1}^{t-1} \left\| \prod_{j=k}^{t-1} \mathbf{W}^{(j)}
\left(\mathbf{I} - \frac{\mathbf{11}^\top}{N}\right) \right\|_F^2   \notag
=\sum_{l=1}^{B-2} (1-p)^{l} \leq \frac{1-p}{p} \notag
\end{align}

After applying the introduced decomposition and separating the product terms,
\begin{align}
  \sum_{k=t-B+1}^{t-1} \left\| \mathbf{G}^{(k)} \prod_{j=k}^{t-1} \mathbf{W}^{(j)}
\left(\mathbf{I} - \frac{\mathbf{11}^\top}{N}\right) \right\|_F^2  \leq \frac{1-p}{p}\hat{\tau} \notag
\end{align}

Therefore, we derive the following.

\begin{align}
   T_{4}\leq \,\hat{\tau}\frac{(1-p)}{p}
+ \left\| \mathbf{G}^{(t-B)} \boldsymbol{\psi}^{(B)} \left(\mathbf{I} - \frac{\mathbf{11}^\top}{N}\right) \right\|_F^2  \notag
\end{align}

Using Corollary \ref{th:corol2} and Assumption \ref{assump:heter},

\begin{align}
    T_5 &= \left\| \mathbf{G}^{(t-B)} \boldsymbol{\psi}^{(B)} \left(\mathbf{I} - \frac{\mathbf{11}^\top}{N}\right) \right\|_F^2     \notag
\\&= \left\| \mathbf{G}^{(t-B)} - \bar{g}^{(t-B)} 1^\top \right\|_F^2
\left\| \boldsymbol{\psi}^{(B)} - \frac{\mathbf{11}^\top}{N} \right\|_F^2     \notag
\\& \leq \hat{\tau} \, C^2 N^2 \lambda^{2B}   \notag
\end{align}

Therefore, we can derive the following.

\begin{align}
    &\|\mathbf{X}^{(t)} - \bar{\mathbf{X}}^{(t)}\|_F^2
\leq \left(1 - \frac{p}{2}\right)\|\mathbf{X}^{(t-B)} - \bar{\mathbf{X}}^{(t-B)}\|_F^2 \notag
 \\&+\left(1 + \frac{2}{p}\right)\eta^2 \hat{\tau}\big(\frac{(1-p)}{p} + C^2 N^2 \lambda^{2B}\big)       \notag
\end{align}

\begin{align}
    &\|\mathbf{X}^{(t)} - \bar{\mathbf{X}}^{(t)}\|_F^2    \notag
\leq \left(1 - \frac{p}{2}\right)^{\tfrac{t-1}{B}} \|\mathbf{X}^{(0)} - \bar{\mathbf{X}}^{(0)}\|_F^2
\\&+ \left(1 + \frac{2}{p}\right)\eta^2 \hat{\tau}\Big(\frac{(1-p)}{p} + C^2 N^2 \lambda^{2B}\Big)   \notag
\sum_{j=0}^{\tfrac{t-1}{B}} \left(1 - \frac{p}{2}\right)^j
\end{align}

Consequently, we can prove the lemma and show the following.

\begin{align}
   &\|\mathbf{X}^{(t)} - \bar{\mathbf{X}}^{(t)}\|_F^2 \leq \notag
   \\& \left(1 + \frac{2}{p}\right)\frac{2}{p}\,\eta^2 \hat{\tau}\Big(\frac{(1-p)}{p} + C^2 N^2 \lambda^{2B}\Big)   \notag
\end{align}

\subsection{Needed Properties}\label{prop:properties}
\textbf{Property 1 (L-Smoothness and Convexity)} \\
Assuming convexity, we have:

\begin{align}
    \langle \nabla f_i(\tilde{\mathbf{x}}), \tilde{\mathbf{x}} - \mathbf{x} \rangle \geq f_i(\tilde{\mathbf{x}}) - f_i(\mathbf{x}). \notag
\end{align}

Under L-Smoothness:

\begin{align}
    F_i(\mathbf{x}, \xi) \leq F_i(\tilde{\mathbf{x}}, \xi) + \langle \nabla F_i(\tilde{\mathbf{x}}, \xi), \mathbf{x} - \tilde{\mathbf{x}} \rangle
    + \frac{L}{2} \|\mathbf{x} - \tilde{\mathbf{x}}\|_2^2. \notag
    \quad \\
    f_i(\mathbf{x}) \leq f_i(\tilde{\mathbf{x}}) + \langle \nabla f(\tilde{\mathbf{x}}), \mathbf{x} - \tilde{\mathbf{x}} \rangle \notag
    + \frac{L}{2} \|\mathbf{x} - \tilde{\mathbf{x}}\|_2^2. 
    \quad
\end{align}

Under the convexity of $F_i$' s, we have:

\begin{align}
    &\|\nabla f_i(\mathbf{x}) - \nabla f_i(\tilde{\mathbf{x}})\|_2 \leq L \|\mathbf{x} - \tilde{\mathbf{x}}\|_2, \notag \\&
    \|\nabla f_i(\mathbf{x}) - \nabla f_i(\tilde{\mathbf{x}})\|_2^2 \leq 
    2L\Big(f_i(\mathbf{x}) - f_i(\tilde{\mathbf{x}}) - \langle \nabla f_i(\tilde{\mathbf{x}}), \mathbf{x} - \tilde{\mathbf{x}}\rangle\Big), \notag
     \\&\|\nabla F_i(\mathbf{x}, \xi) - \nabla F_i(\tilde{\mathbf{x}}, \xi)\|_2^2 \leq  \notag
    \\&2L\Big(F_i(\mathbf{x}, \xi) - F_i(\tilde{\mathbf{x}}, \xi) - \langle \nabla F_i(\tilde{\mathbf{x}}, \xi), \mathbf{x} - \tilde{\mathbf{x}}\rangle\Big). \notag
\end{align}

\textbf{Property 2 (Norm inequalities)}
\begin{itemize}
    \item \textbf{$\mathbf{a}_i \in \mathbb{R}^d$:}
    \[
    \left\| \sum_{i=1}^n \mathbf{a}_i \right\|_2^2 \leq n \sum_{i=1}^n \|\mathbf{a}_i\|_2^2.
    \]

    \item \textbf{$a, b \in \mathbb{R}^d$:}
    \[
    \|\mathbf{a}+\mathbf{b}\|_2^2 \leq (1+\alpha)\|\mathbf{a}\|_2^2 + (1+\alpha^{-1})\|\mathbf{b}\|_2^2, 
    \quad \forall \alpha > 0.
    \]
\end{itemize}

\textbf{Property 3.}

It is obvious that the equation below holds due to the doubly stochasticity of $\mathbf{W^{(t)}}$.

\begin{align}
    & (\mathbf{W}^{(t-2)}\mathbf{W}^{(t-1)}- \frac{\mathbf{11}^\top}{N}) = \notag \\ &(\mathbf{W}^{(t-2)}- \frac{\mathbf{11}^\top}{N})(\mathbf{W}^{(t-1)}- \frac{11^\top}{N}) \notag
\end{align}

A Similar decomposition can be applied to any term with the format of $(\prod_{j=k}^{t-1}\mathbf{W}^{j}- \frac{\mathbf{11}^\top}{N})$. Therefore, for a fixed k, we can derive:

\begin{align}
    \left\| \prod_{j=k}^{t-1} \mathbf{W}^{(j)}
\left(\mathbf{I} - \frac{\mathbf{11}^\top}{N}\right) \right\|_F^2 =& \prod_{j=k}^{t-1} \left\| \mathbf{W}^{(j)} - \frac{\mathbf{11}^\top}{N} \right\|_F^2 \notag
\\
\leq & (1-p)^{t-1-k}. \notag
\end{align}

where $t-1-k$ shows the number of mixing matrices in the product term.

\newpage

\section{Biography Section} 

% \end{IEEEbiography}
\vspace{11pt}

\vfill

\end{document}